\documentclass{article} 
\usepackage{iclr2027_conference,times}

\usepackage{amsmath,amsfonts,bm}

\def\eqref#1{equation~\ref{#1}}

\def\1{\bm{1}}

\DeclareMathAlphabet{\mathsfit}{\encodingdefault}{\sfdefault}{m}{sl}
\SetMathAlphabet{\mathsfit}{bold}{\encodingdefault}{\sfdefault}{bx}{n}

\usepackage{hyperref}
\usepackage{url}
\usepackage{xspace}
\usepackage{graphicx}
\usepackage{booktabs}
\usepackage{array}
\usepackage{float}
\usepackage{subcaption}
\usepackage{xurl}
\hypersetup{hidelinks}

\newcommand{\ourmethod}{\textsc{TMax}\xspace}
\newcommand{\ourdata}{\textsc{Tmax-13k}\xspace}
\newcommand{\ourmodel}{\textsc{Tmax-9B}\xspace}
\newcommand{\ourfourbmodel}{\textsc{Tmax-4B}\xspace}
\newcommand{\ourtwobmodel}{\textsc{Tmax-2B}\xspace}
\newcommand{\ourtwentysevenbmodel}{\textsc{Tmax-27B}\xspace}

\title{\ourmethod: A simple recipe for terminal agents}

\newcommand{\samethanks}[1][\value{footnote}]{\footnotemark[#1]}
\author{
  \textbf{Hamish Ivison\thanks{Equal contribution. Work done while HI and NL were at Ai2.}}\hspace{0.4em}$^{\,\alpha\,\omega}$ \hspace{1.2em}
  \textbf{Junjie Oscar Yin\samethanks}\hspace{0.4em}$^{\,\omega}$
  \textbf{Rulin Shao\samethanks}\hspace{0.4em}$^{\,\omega}$ \\
  \textbf{Teng Xiao}$^{\,\alpha}$ \quad
  \textbf{Nathan Lambert}$^{\,\alpha}$ \quad
  \textbf{Hannaneh Hajishirzi}$^{\,\omega}$ \\[4pt]
  $^{\alpha}$Allen Institute for AI \quad $^{\omega}$University of Washington \\
  \texttt{\{hamishiv,osey\}@cs.washington.edu}
}

\iclrfinalcopy 
\begin{document}

\maketitle
\lhead{Preprint}

\begin{abstract}
Terminal-using agents have quickly become a popular and important downstream application of language models (LMs). Despite their prevalence, relatively little academic work has examined RL-based training of these models, due to difficult benchmarks, a lack of data, and a lack of simple baseline recipes. 
We present \ourmethod, an open RL recipe for terminal agents that brings small open models closer to the frontier.
While simple, our recipe achieves \textbf{33\% on Terminal-Bench 2.0 with only 9B parameters}, outperforming much larger models from prior work.
Our recipe first involves generating data using a novel taxonomy, combining difficulty control, personas, and verifier diversification, which allows us to cheaply generate large amounts of terminal environments for RL and SFT training.
We then train open-weight models using RL with our data, using a simple, outcome-only recipe, with small adjustments to improve stability.
Our recipe works for models from 2B to 27B and across model families, dominating the Pareto curve for open terminal-agent performance at smaller sizes.
We release our data, models, and code as a strong baseline for future open academic work on terminal agents at \url{https://github.com/hamishivi/tmax}.
\end{abstract}

\begin{figure}[h]
  \centering
  \includegraphics[width=\textwidth]{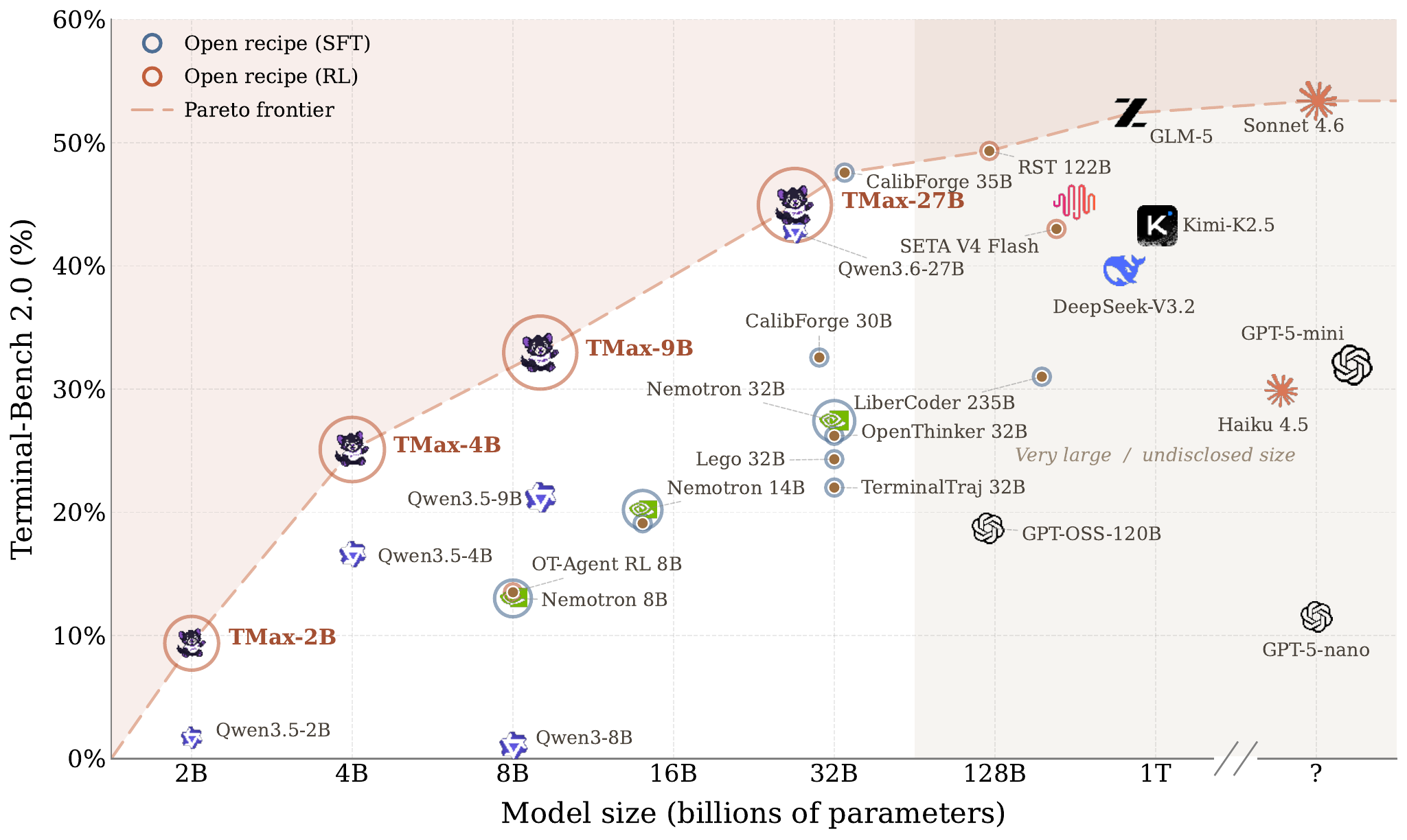}
  \caption{Terminal-Bench 2.0 performance of \ourmethod alongside prior and concurrent models. \ourmethod and Qwen use our harness; other scores are self-reported or from the leaderboard.
  \textbf{\ourmodel dominates the Pareto curve for models under 10B parameters.}
  See Table~\ref{tab:tb2-teaser_table} for full results.}
  \label{fig:teaser}
\end{figure}

\section{Introduction}

Terminal-based agentic coding products have quickly become incredibly popular, with models expected to perform increasingly complex and long-running tasks through the terminal~\citep{anthropic_claude_code, research2026composer2technicalreport}.
However, existing academic work has largely focused on bug-fixing~\citep{jimenez2024swebench} or relatively simple terminal tasks~\citep{openthoughts-agent,gandhi2025endless}, rather than the complex, long-horizon tasks exemplified by Terminal-Bench~\citep{merrill2026terminal}.
We aim to close this gap by providing a large dataset of complex terminal tasks and a recipe for training small yet capable open-weight terminal agents.

We first introduce \ourdata{}, a dataset of RL environments spanning
diverse domains, skills, and difficulty levels.
We explicitly control task difficulty and use continuous rewards
where appropriate, while keeping the generation process simple and scalable.
Compared with many prior datasets, \ourdata{} is substantially
more challenging, as measured by lower pass rates from
Gemini-3-Flash-Preview. 
While we get the strongest results using stronger models as data generators, we also show that training on data generated by the same model can still provide strong improvements, showing that our pipeline is effective in the absence of a stronger teacher model.
Secondly, while recent work has focused primarily on generating terminal-agent data~\citep{wu2026termitraj,zhu2026termigen}, RL training for terminal agents remains underexplored. We address this gap by developing a simple RL recipe for training
open-weight terminal agents on \ourdata{}.
Our recipe builds on fully asynchronous RL infrastructure implemented
in open-instruct~\citep{lambert2025tulu3pushingfrontiers} and uses
DPPO~\citep{qi2026rethinkingtrustregionllm} for policy optimization.
To improve training stability, we use an FP32 LM head and increase
the group size.
\textbf{Our best 9B model achieves 33\% on Terminal-Bench 2.0},
outperforming similarly sized open models and selected closed models,
including Claude Haiku 4.5.


We hope that our overall recipe and data prove useful for future academic work exploring terminal agents, as we believe this setting provides a number of unique and interesting challenges.
Concretely, our contributions are:
\begin{itemize}
    \item We introduce \ourdata, a dataset of 13,565 RL environment instances with explicit control over difficulty and diversity. Our data is generated using a simple and scalable procedure, avoiding complex multi-agent loops.
    \item We train open-weight terminal agents and achieve state-of-the-art performance among reported models under 10B parameters under default Terminal-Bench settings. Our best 9B model reaches 33\% on Terminal-Bench 2.0.
    \item  We provide a simple, reproducible open RL recipe for achieving this performance and publicly release all elements required for training our models. Our recipe outperforms prior open RL recipes such as Endless Terminals and OpenThinker-Agent.
    \item We show that terminal-based RL training can generalise non-trivially across harnesses and tasks, providing strong evidence our training improves general model capabilities.
\end{itemize}

We additionally publicly release code, checkpoints and data associated with this work for future use and reproduction at \url{https://github.com/hamishivi/tmax}.

\vspace{-0.5em}
\section{Background and Related Work}
\vspace{-0.25em}
\paragraph{Data for terminal agents.}
Early terminal-agent datasets paired short instructions with shell
commands~\citep{lin-etal-2018-nl2bash}. More recent work either
constructs environments from existing repositories or
questions~\citep{yang2025swesmith,wu2026termitraj,
lin2026cligymscalableclitask,yang2026terminallego,shen2026seta,shi2026facet},
or synthesises them using seed tasks and combinatorial
prompting~\citep{gandhi2025endless,zhu2026termigen,
peng2026litecoderterminal,li2026recursivesynthesislonghorizonterminal,
meng2026calibforge}. We follow the latter approach. We find that
repository-based construction often produces imbalanced domain
coverage, whereas synthesis provides greater control over domain
balance. Compared with prior work on task synthesis, we vary tasks
along more axes, including task difficulty, and release substantially
more data. We also incorporate a repair stage and show that strong
performance can be achieved without a commonly used but expensive
validation step. We discuss prior and concurrent work on terminal
RL data in more detail in \S~\ref{app:related_work}.
\vspace{-0.25em}
\paragraph{RL for terminal agents.}
Closest to our work is Endless Terminals~\citep{gandhi2025endless}, which performs RL training
on a Qwen 3 8B model using its generated data, but with a significantly simpler harness and setup.
OpenThoughts-Agent~\citep{raoof2026openthoughtsagent} explores RL training for terminal agents, but focuses on comparing existing datasets and the effect of varied SFT, finding that RL dataset quality matters significantly.
More recently, SETA~\citep{shen2026seta} and ROME~\citep{wang2026letflowagenticcrafting} explore terminal RL training, but show limited gains or use fairly complex setups to train their models. In contrast, we show large gains and use a simple, straightforward recipe. River~\citep{yao2026river} examines RL training data, finding that rubric-based repair and filtering of existing datasets can improve performance. It also investigates process-level behaviour regularisation. In contrast, we generate our own data, and use fairly straightforward outcome rewards.

\begin{figure*}[t]
    \centering
    \resizebox{1.0\textwidth}{!}{%
    \includegraphics{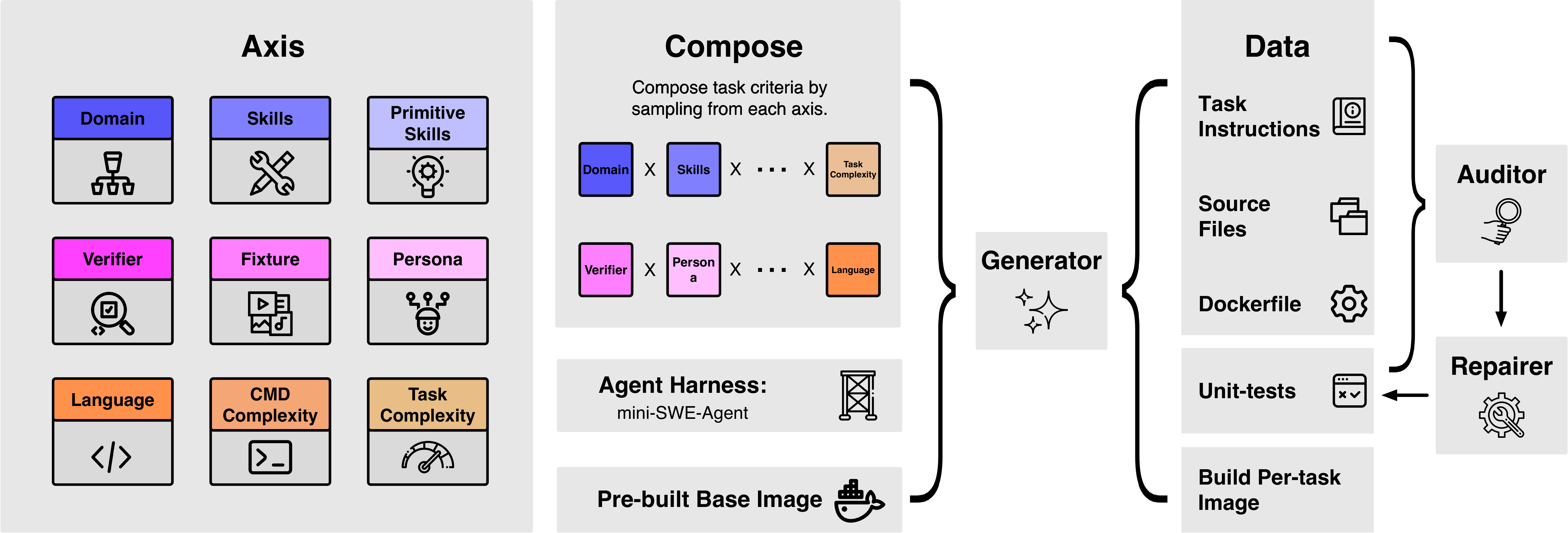} 
    }
    \caption{\textbf{\ourmethod Data Pipeline}. Each task is composed by hierarchically sampling from 9 structured axes, after which a \emph{generator} instantiates a task as a Dockerfile, unit-test verifier, source files, and task instructions. An \emph{auditor} then labels whether each verifier actually validates its instruction---flagging answer leakage, instruction--verifier mismatch, and overly weak checks---and a \emph{repairer} rewrites the verifier where fixable. Unrepairable tasks are dropped. Tasks are built atop a pre-built per-domain base image
  and served through a mini-SWE-agent harness. Composing axes yields combinatorially many task signatures with explicit, per-axis control over difficulty and diversity. The single build step, with audit and repair as a one-shot post-pass, keeps generation cheap
  to scale.}
    \label{fig:pipeline}
\end{figure*}

\section{Terminal Data Generation}
Training autonomous terminal agents requires large-scale and diverse coding tasks. With real-world terminal data often scarce or proprietary \citep{merrill2026terminal}, synthetic generation is used to bridge the data gap \citep{pi2026data, zhu2026termigen}.
However, existing pipelines fall short along three axes. First, they rely on complex multi-stage generation procedures that are hard to scale. Second, they generate homogeneous task suites with limited coverage against real-world tasks and Terminal-Bench \citep{merrill2026terminal}. Third, they offer little control over difficulty, often producing bi-modal task pools that are either trivially solved or unsolvable by current models. 
Hence, we propose a simple terminal environment generation pipeline that is scalable, diverse, and difficulty-aware. We use Gemini-3-Pro \citep{pichai2025new} as our generation model, as it requires a strong enough model to generate coherent tasks and code fixtures.

\subsection{Generation Pipeline}
\label{sec:gen_pipeline}
We adopt a simple compositional generation framework: each synthetic task is sampled as a product of structured axes (Table~\ref{tab:axes}). We follow \citet{pi2026data} to seed the first three axes -- domains, skills, primitive skills -- and introduce six orthogonal axes explicitly targeting diversity and difficulty, unlike prior work.

\paragraph{Diversity via hierarchical sampling.} The compositional sampler is itself our main diversity mechanism, and drawing from each axis yields combinatorially many distinct task signatures. Further, we add two specific diversity dimensions:

\emph{Persona diversification.} We augment generation with a set of user personas, motivated by prior work on persona-conditioned data synthesis~\citep{ge2025scalingsyntheticdatacreation, lambert2025tulu3pushingfrontiers}, to further diversify generated tasks. Personas are domain-specific (6--18 per domain, e.g.\ ``red-team operator crafting an evasion payload'' for security), so each is always reasonable for its sampled skill set.

\emph{Multi-modal fixtures.} Prior work on RL environments largely focuses on \emph{text-in / text-out} problems. Here we extend the task inputs beyond plain text by shipping a concrete artifact for some tasks: a PNG image, audio file, video, stripped binary, vendored package, or multi-service compose stack (Table~\ref{tab:fixture-kinds}). The policy itself remains a text-only language model and training is unchanged: rather than perceiving these artifacts natively, the agent inspects them through standard terminal tooling (e.g., OCR, audio transcription, \texttt{ffmpeg}), so no multi-modal model is required.

\paragraph{Difficulty via explicit calibration.} Unlike previous works, we make an explicit effort to calibrate task difficulty, avoiding the bi-modal distribution where tasks are either trivially solved or unsolvable by current models (Table~\ref{tab:terminal-data-difficulty}). We address this through two mechanisms:

\emph{Fine-grained complexity.} We introduce two complexity axes, \emph{task complexity} (from a few shell commands to intricate workflows of 30--60 commands) and \emph{command complexity} (from bash-only to bash + code + system services), for granular control over how hard each task is. We sample uniformly across complexity buckets by default, with optional per-bucket up-weights to match specific model capabilities or induce a curriculum.

\emph{Graded verifiers.} Previous RL tasks often rely on exactly matching a gold output. Here we extend with richer verifier kinds (Table~\ref{tab:verifier-kinds}): metric-threshold (e.g., accuracy $\geq 0.95$), adversarial-corpus (accept clean, reject malicious), fuzz-equivalence (bit-exact match against an oracle), or multi-protocol (protocol-level requests against a service). Thresholds give us a continuous difficulty knob, while multi-condition variants naturally extend task length.

\paragraph{Verification via LM Judge.} Finally, inspired by \citet{yao2026river}, we apply an LM judge (GPT 5.6 Sol) to classify our data and use it to identify weak, ambiguous, or contradictory verifiers and task descriptions.
The vast majority (76.4\%) of tasks are marked as having verifiers too weak to enforce the correct solution, and 10.9\% of tasks are judged as leaking the answer in the initial environment state, highlighting that a repair pass is much-needed to improve task validity.
We then provide the original environment definition along with the label to GPT 5.6 Sol and construct a repaired instance. We then do a final spot-check to ensure that images have the required dependencies and can be built (see \S\ref{app:repair_stats} for the pre-repair rubric counts). We retain tasks originally judged as correct and repaired tasks.

\paragraph{Do we need to filter?} Unlike prior works that validate task quality and correctness through expensive teacher generation~\citep{zhu2026termigen, wu2026termitraj}, we deliberately skip this step entirely, as our RL training applies effective soft filtering via active sampling and zero-std filtering~(\S\ref{sec:RL-Method}).
We find the all-zero rate is roughly 40\% (see \S\ref{app:filtered_samples}).

Using this pipeline, we generate a dataset of 13,565 tasks, which we name \ourdata. We measure overlap between dataset task descriptions and the Terminal-Bench and TB Lite tasks using a 13-gram sliding window, and find no sign of overlap (\S\ref{app:decontamination}).



\begin{figure*}[t]
    \centering
    \includegraphics[width=.8\textwidth]{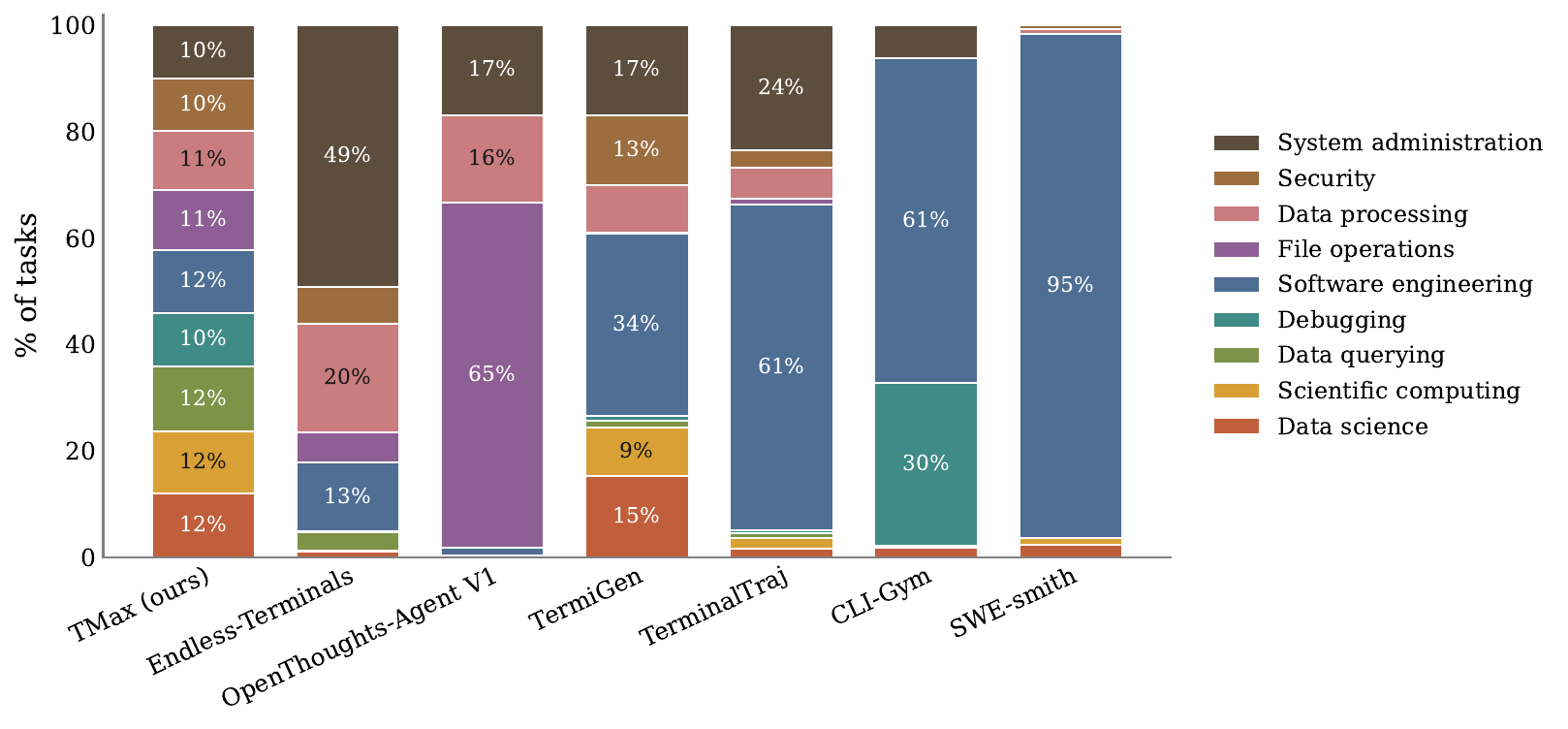}
    \caption{\textbf{Data Composition}. Domain distribution of tasks across terminal datasets. Prior datasets skew heavily toward one or two domains, whereas our compositional sampler yields balanced coverage across all nine domains. In-bar percentages are rounded to the nearest whole number. OpenThoughts Agent refers to the original OpenThoughts data~\citep{openthoughts-agent}.}
    \label{fig:composition}
\end{figure*}

\subsection{Comparing Terminal Datasets}

We compare our generated data against past work, both against a subset of prior terminal agent datasets, and a prior SWE-bench-targeting dataset, SWE-Smith~\citep{yang2025swesmith}.

\begin{table*}[t]
\centering
\vspace{0.2em}
\resizebox{0.95\textwidth}{!}{%
\begin{tabular}{@{\hspace{0.4em}}l r c c c c c c c@{\hspace{0.4em}}}
\toprule
\textbf{Data} &
\textbf{Dataset size} &
\shortstack{\textbf{Pass@1}\\\textbf{Gemini}} &
\shortstack{\textbf{Pass@4}\\\textbf{Gemini}} &
\shortstack{\textbf{Pass@8}\\\textbf{Gemini}} &
\shortstack{\textbf{Mean}\\\textbf{turns}} &
\shortstack{\textbf{Mean tokens}\\\textbf{/ run}} &
\shortstack{\textbf{Domain}\\\textbf{balance}} &
\shortstack{\textbf{Skill-type}\\\textbf{balance}} \\
\midrule
\textbf{\ourdata (Ours)} & 13k & 49\% & 57\% & 60\% & 31.5 & 446K & 0.997 & 0.738 \\
\addlinespace[0.8ex]
\cmidrule(lr){1-9}
Endless Terminals & 2.4k & 92\% & 94\% & 95\% & 15.7 & 117K & 0.481 & 0.284 \\
Terminal Gen & 3k & 57\% & 64\% & 66\% & 20.9 & 305K & 0.646 & 0.477 \\
Terminal Traj & 5.5k & 54\% & 63\% & 65\% & 15.1 & 141K & 0.363 & 0.374 \\
CLI-Gym & 1.5k & 41\% & 52\% & 55\% & 43.8 & 769K & 0.283 & 0.061 \\
SWE-Smith & 59k & 54\% & 69\% & 72\% & 41.2 & 582K & 0.146 & 0.042 \\
\bottomrule
\end{tabular}%
}
\caption{\textbf{Terminal datasets and difficulty-related statistics} with Gemini-3-Flash-Preview, evaluated on a fixed subsample of 250 tasks per dataset with 8 rollouts each. Pass@$k$ is mean pass@$k$ across tasks. Mean tokens/run is the sum of tokens over turns, in thousands. For domain and skill-type balance, higher indicates more even category coverage with a maximum of 1.}
\vspace{-0.1in}
\label{tab:terminal-data-difficulty}
\end{table*}

\paragraph{Composition} A good training dataset should have broad coverage and a balanced composition across target domains. By design, our generation framework exposes domain as an explicit sampling axis, so we can directly calibrate per-domain mass at generation time. To compare across datasets, we use Gemini-3-Pro to annotate every task for our and six comparison datasets, with one of these nine domain labels. As shown in Figure~\ref{fig:composition}, prior terminal datasets concentrate 34--95\% of their mass on a single domain (e.g., \texttt{software\_engineering} accounts for 95\% of SWE-Smith and over 60\% of TerminalTraj and CLI-Gym), while ours spreads roughly uniformly across all nine. This balanced coverage avoids biasing the policy toward any narrow domain and better covers real-world usage.

\paragraph{Difficulty \& Balance}

To compare difficulty across datasets, we evaluate Gemini-3-Flash-Preview on a fixed subsample of 250 tasks per dataset (8 rollouts each) and report mean pass@$k$ in Table~\ref{tab:terminal-data-difficulty}. We additionally compute a \emph{balance score} for each categorical axis (domain, skill-type, task complexity, command complexity) as the fraction-of-uniform diversity of its empirical distribution:
\begin{equation}
\label{eq:balance}
    \mathrm{Balance} \;=\; \frac{\exp(H)}{N}, \qquad H = -\sum_{i=1}^{N} p_i \log p_i,
\end{equation}
where $p_i$ is the proportion of tasks in bucket $i$ and $N$ is the number of buckets. The score lies in $[1/N, 1]$ with $1$ being perfectly uniform; see \S\ref{app:balance} for derivation.
This highlights that our data requires significantly more turns than other datasets apart from CLI-Gym and SWE-Smith, but is significantly more balanced than them. Difficulty-wise, we observe that it has the lowest pass@8 apart from CLI-Gym, suggesting genuine difficulty and not high variance in rollout outcomes.
While some of these tasks may be unfeasible for the model, we do not attempt to filter them out ahead of time as our RL infrastructure already filters out zero pass rate instances.

\section{Training Terminal Agents}

We now validate our data by practically applying it to RL training. We aim to validate our data with a simple recipe, providing a testbed in which we believe there is significant room for improvement.

\subsection{Experimental Setup}
\label{sec:RL-Method}

\paragraph{Algorithm}

We train models using DPPO~\citep{qi2026rethinkingtrustregionllm}, a variant of GRPO~\citep{shao2024deepseekmathpushinglimitsmathematical} that masks tokens when inference and training logprobs deviate. Specifically, we mask tokens based on a binary approximation of the total variation (TV) divergence. We use a token level loss~\citep{yu2025dapoopensourcellmreinforcement} and run training fully asynchronously. We filter groups with zero standard deviation and use active sampling to ensure full batches following Olmo 3~\citep{olmo2026olmo3}.

\paragraph{Infrastructure}
We extend open-instruct~\citep{lambert2025tulu3pushingfrontiers} for terminal agent training. For rollouts, we use vLLM~\citep{kwon2023efficient}, and we use the same mini-SWE-agent-based harness as used for verification and SFT data generation.
We build a custom sandboxing infrastructure utilising slurm for requesting additional compute resources, and Apptainer for setting up and isolating environments.
We additionally ensure that the language model head of models is kept in FP32 precision to minimise training-inference mismatch \citep{minimax2025minimaxm1scalingtesttimecompute} (see~\S\ref{app:lm_head}).
We train on a cluster of H100 nodes, and use 2 nodes for training and 6 for inference. Training typically takes 1--2 days.
We remove a small number of training data samples and eval samples that are unable to run correctly in our infrastructure (see \S\ref{app:infra_filter}).

\paragraph{Harness.} Unless otherwise stated, we use a simple harness based on mini-SWE-agent \citep{yang2024swe} with persistent shell. We find that Terminus-2 harness \citep{merrill2026terminal} is more brittle with small models, as it requires agents to send raw keystrokes to interact with the terminal (see \S\ref{app:harness_choices}).

\paragraph{Evaluation} We primarily evaluate on Terminal-Bench 2.1~\citep{merrill2026terminal} and Terminal-Bench Lite~\citep{OpenThoughts-TBLite}. 
We use Harbor~\citep{Harbor_Framework} and our custom sandboxing infrastructure for evaluation. 
Due to compatibility issues, we filter a small number of samples from Terminal-Bench Lite and Terminal-Bench 2.1 which are unsupported by our sandbox (see \S\ref{app:eval_filtering}). 
We run 3 trials for all evaluations and report mean and standard error. We run all evals using a single H200 and 4 concurrent trials. We do not override timeout defaults.

For the evaluations in Fig.~\ref{fig:teaser}, we use Terminal-Bench 2.0 (the earlier version) to compare directly with past work. We additionally use Daytona as the backend as recommended by Terminal-Bench authors to make results more reproducible. We find that the choice of backend can non-trivially affect performance: for instance, Daytona sandboxes often perform installs faster than our locally-hosted runs, resulting in fewer timeouts. However, the cost of these services makes them expensive to develop against.\footnote{One RL training run of \ourmodel would cost roughly \$3,150 on Daytona, using average runtimes from the run and pricing from \url{https://www.daytona.io/pricing}.}

\paragraph{Hyperparameters} Unless otherwise stated, we run RL training for 500 steps and evaluate every 100 steps. For all runs, we report scores on the checkpoint scoring the highest on Terminal-Bench Lite to avoid unfairly biasing against some datasets that collapse later during training in our setup (e.g., CLI-Gym).
We use a group size of 32 and 8 prompts per batch, with a maximum sequence length of 65536 tokens and 64 maximum tool calls. When training Qwen 3.5 and 3.6 models, we do not perform any initial SFT. We edit chat templates such that thinking is preserved on intermediate turns (often called `interleaved thinking'), which has been shown to improve performance in agentic settings~\citep{minimax2025interleaved}. We show further hyperparameters in \S\ref{app:full_rl_hypers}.

\subsection{Core Results}
\label{sec:core_results}
\begin{table*}[t]
\centering
\small
\begin{tabular}{lcc}
\toprule
RL Dataset / Model & TB Lite & TB 2.1 \\
\midrule
None (Qwen3.5-9B) & $49.4 \pm 1.9$ & $20.5 \pm 1.7$ \\
\midrule
TerminalTraj~\citep{wu2026termitraj} & $52.5 \pm 1.9$ & $15.9 \pm 1.1$ \\
TermiGen~\citep{zhu2026termigen} & $54.7 \pm 2.0$ & $21.2 \pm 1.2$ \\
LiteCoder~\citep{peng2026litecoderterminal} & $56.5 \pm 1.8$ & $22.0 \pm 1.4$ \\
OpenThinker-Agent~\citep{raoof2026openthoughtsagent} & $59.4 \pm 2.1$ & $22.7 \pm 1.6$ \\
CLI-Gym~\citep{lin2026cligymscalableclitask} & $61.6 \pm 1.9$ & $24.6 \pm 1.6$ \\
SETA~\citep{shen2026seta} & $59.2 \pm 1.8$ & $26.5 \pm 1.8$ \\
CalibForge~\citep{meng2026calibforge} & $59.9 \pm 2.1$ & $26.5 \pm 1.7$ \\
Terminal-Lego~\citep{yang2026terminallego} & $63.2 \pm 2.0$ & $26.5 \pm 1.8$ \\
RST~\citep{li2026recursivesynthesislonghorizonterminal} & $63.4 \pm 1.8$ & $27.3 \pm 1.6$ \\
SWE-Smith~\citep{yang2025swesmith} & $60.2 \pm 1.9$ & $27.3 \pm 1.5$ \\
Endless Terminals~\citep{gandhi2025endless} & $62.0 \pm 1.9$ & $28.4 \pm 1.6$ \\
Facet~\citep{shi2026facet} & $61.3 \pm 2.1$ & $28.8 \pm 1.6$ \\
\midrule
\textbf{\ourdata (Ours)} & $\boldsymbol{65.6 \pm 2.0}$ & $\boldsymbol{29.6 \pm 1.8}$ \\
\bottomrule
\end{tabular}
\caption{\textbf{Comparing different RL datasets on Terminal-Bench.} We train on varied prior and concurrent RL datasets using our recipe and evaluate on Terminal-Bench Lite and 2.1. We find that \ourdata achieves the highest mean score on both evaluations.}
\label{tab:termbench}
\end{table*}

\paragraph{\ourdata outperforms other terminal datasets.} We first apply RL training to Qwen 3.5 9B~\citep{qwen3.5} using a variety of datasets from prior work and show results in Table~\ref{tab:termbench}. 
We include prior datasets focusing on Terminal-Bench alongside SWE-Smith, as a representative sample of a dataset focusing on SWE-Bench tasks. We note that in many cases, we are the first to openly apply these datasets for RL training (e.g., in the case of TermiGen and TerminalTraj). We call the model trained on our data \ourmodel. We find that training on \ourdata matches or exceeds training on other datasets, and achieves the highest mean performance across both datasets.

\begin{figure}
    \centering
    \begin{minipage}{.49\textwidth}
        \centering
        \includegraphics[width=\linewidth]{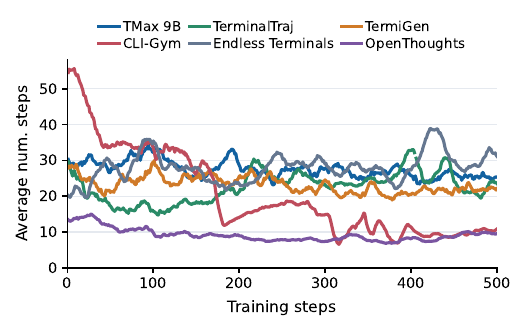}
        \caption{Average step count over RL training when training Qwen 3.5 on different datasets. We smooth using a 15-step average window. \textbf{Training on \ourdata consistently uses higher steps throughout training than other datasets}, apart from Endless Terminals.}
        \label{fig:step_count_trainer}
    \end{minipage}\hfill
    \begin{minipage}{.49\textwidth}
        \centering
        \includegraphics[width=.83\linewidth]{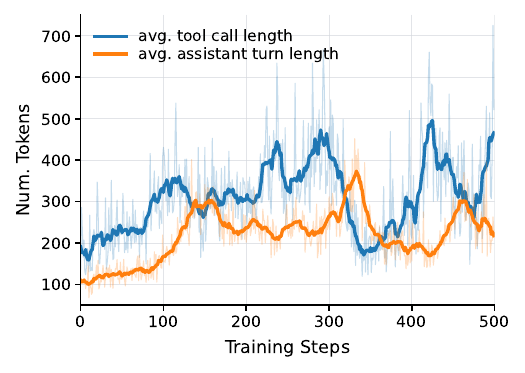}
        \caption{Average length (in tokens) of assistant turns and tool calls. We do not include tool calls in the assistant turn length. Over the initial 150 training steps, \textbf{model output lengths increase alongside its performance}.}
        \label{fig:tool_call_tokens}
    \end{minipage}
\end{figure}

To further investigate this, we plot the average number of steps taken by the model during training for each dataset in Fig.~\ref{fig:step_count_trainer}. We find that some lower scoring datasets (OpenThoughts) use fewer turns during training or even collapse in turn count over the course of training, indicating lower difficulty. In contrast, \ourdata remains stable at requiring 25-30 steps to solve, maintaining high difficulty throughout training.
While easier datasets like Endless Terminals also maintain higher step counts, they do so at the cost of filtering increasingly more fully-solved samples during training (see \S\ref{app:filtered_samples}), requiring up to seven times as many rollouts as the minimum needed to produce a single batch.

We additionally plot reasoning and tool-call token usage in Fig.~\ref{fig:tool_call_tokens}. We observe token usage increasing over the initial 150 training steps before plateauing (and then becoming somewhat unstable later in training).
This is reminiscent of inference-time reasoning scaling in single-turn math settings~\citep{Guo_2025}, and suggests that the model is gradually learning more complex reasoning and more complex tool calls over the course of training, resulting in its improved performance. 

\paragraph{\ourmodel outperforms open RL recipes on Terminal-Bench 2.0.} 
We additionally compare our model to prior work's self-reported scores on Terminal-Bench 2.0 in Fig.~\ref{fig:teaser}.
This shows that even when comparing across different harnesses and training setups, \ourmodel performs the strongest among models under 10B parameters.
In particular, our model outperforms many past open RL recipes such as Endless Terminals~\citep{gandhi2025endless} and OpenThoughts-Agent~\citep{raoof2026openthoughtsagent}.
As such, our work provides a strong baseline for future open work on agentic terminal RL.

\begin{table}[t]
\centering
\begin{tabular}{lcc}
\toprule
\textbf{Model} & \textbf{TB Lite} & \textbf{TB 2.1} \\
\midrule
Qwen 3.5 2B & $7.1 \pm 1.0$ & $1.9 \pm 0.4$ \\
\quad + RL (\ourtwobmodel) & $\boldsymbol{18.2 \pm 1.4}$ & $\boldsymbol{5.3 \pm 1.0}$ \\
\midrule
Qwen 3.5 4B & $33.6 \pm 1.8$ & $15.5 \pm 1.5$ \\
\quad + RL (\ourfourbmodel) & $\boldsymbol{54.3 \pm 1.9}$ & $\boldsymbol{24.6 \pm 1.8}$ \\
\midrule
Qwen 3.5 9B & $49.4 \pm 1.9$ & $20.5 \pm 1.7$ \\
\quad + RL (\ourmodel) & $\boldsymbol{65.6 \pm 2.0}$ & $\boldsymbol{29.6 \pm 1.8}$ \\
\midrule
Qwen 3.6 27B & $74.2 \pm 1.5$ & $43.6 \pm 1.9$ \\
\quad + RL (\ourtwentysevenbmodel) & $\boldsymbol{77.4 \pm 1.8}$ & $\boldsymbol{43.9 \pm 1.6}$ \\
\midrule
Qwen 3 8B & $5.0 \pm 1.0$ & $1.1 \pm 0.5$ \\
\quad + SFT & $9.1 \pm 1.3$ & $5.3 \pm 1.1$ \\
\quad + RL & $\boldsymbol{17.8 \pm 1.1}$ & $\boldsymbol{6.4 \pm 1.0}$ \\
\bottomrule
\end{tabular}
\caption{\textbf{\ourdata improves models across different sizes and families.}
We train Qwen 3.5 and Qwen 3 models on \ourdata across sizes, and find it almost always improves performance on both Terminal-Bench benchmarks. 
We evaluate Qwen 3 8B models with a 40,720-token max len.}
\label{tab:qwen35-sizes}
\end{table}

\paragraph{\ourmethod RL training improves models across different sizes.}  We additionally apply our recipe, without modification, to the other sizes of Qwen 3.5 (and Qwen 3.6), resulting in \ourtwobmodel, \ourfourbmodel and \ourtwentysevenbmodel. We show evaluation results in Table~\ref{tab:qwen35-sizes}. In all cases, we improve over the baseline in Terminal Bench 2.1 scores, highlighting that \ourdata contains a wide enough range of difficulty levels to provide training signal even at small (2B) model sizes. 
We find the improvement is the smallest (within noise for TB2.1) for the 27B size model, which we attribute to the much-improved performance of Qwen 3.6 relative to the 3.5 series, making it harder to further improve.

\subsection{Terminal Agent Training can Generalise}

We next investigate how well \ourmodel generalises across three important axes: tasks (i.e., what it is asked to do), harnesses (i.e., what tools and prompts it is provided when performing tasks), and model families (i.e., the starting point model).

\begin{table}[t]
\centering
\begin{tabular}{llcc}
\toprule
Model & Harness & SBV & AIME \\
\midrule
Qwen 3.5 9B & None & -- & $67.5 \pm 4.9$ \\
Qwen 3.5 9B & Ours & $46.2 \pm 0.8$ & $72.2 \pm 2.3$ \\
\quad + RL (\ourmethod) & Ours & $\boldsymbol{55.3 \pm 0.7}$ & $\boldsymbol{93.3 \pm 1.5}$ \\
\bottomrule
\end{tabular}
\caption{\textbf{Improved performance generalises across tasks.} Performance of Qwen 3.5 9B on SWE-bench Verified (SBV) and AIME'24/25 with and without our harness.}
\label{tab:qwen35-sbv-aime}
\end{table}

\paragraph{\ourmethod RL training generalises to other tasks.}
We evaluate \ourmodel on SWE-Bench Verified~\citep{jimenez2024swebench} and AIME 2024/2025, and additionally compare to AIME performance in a traditional single-turn no-harness setting in Table~\ref{tab:qwen35-sbv-aime}.
Encouragingly, we find that performance improves across the board by significant margins, suggesting that the model is not simply fitting to the harness and domain, but also learning how to better make use of its terminal tools to solve generic problems.
In particular, we observe that \ourmodel works out how to use the terminal to aid it with geometry problems in AIME, unlike Qwen 3.5 9B.
This aligns with prior work showing that terminal-agent training can improve general model capabilities~\citep{cheng2026computerenvironmentselicitgeneral}.

\begin{table*}[t]
\centering
\begin{tabular}{lcccc}
\toprule
Model & Ours & OpenHands & mini-SWE-agent & Terminus-2 \\
\midrule
Qwen 3.5 9B & $49.4 \pm 1.9$ & $29.0 \pm 1.6$ & $44.9 \pm 2.0$ & $37.9 \pm 2.2$ \\
\ourmodel & $\boldsymbol{65.6 \pm 2.0}$ & $\boldsymbol{42.7 \pm 1.7}$ & $\boldsymbol{57.8 \pm 2.0}$ & $\boldsymbol{46.1 \pm 2.1}$ \\
\bottomrule
\end{tabular}
\caption{\textbf{Terminal-Bench Lite performance across harnesses.} We find that \ourmodel improves over its base even when evaluated using different harnesses from training.}
\label{tab:qwen35-harness-tblite}
\end{table*}

\paragraph{\ourmethod RL training generalises to other harnesses.} We evaluate Qwen 3.5 9B and \ourmodel on Terminal-Bench Lite using varied harnesses in Table~\ref{tab:qwen35-harness-tblite}.
\ourmodel improves by at least 8 points across all harnesses, although it performs best with our harness. This suggests that single-harness terminal RL training can generalise across other setups, contrary to recent work~\citep{nanorollout}.

\paragraph{\ourmethod RL training improves different model families.} We also apply our recipe to Qwen 3 8B after a warm-start on our SFT data (\S\ref{sec:sft_data_gen}). As seen in Table~\ref{tab:qwen35-sizes}, we still observe Terminal-Bench improvements (especially Terminal-Bench Lite) despite the significantly weaker base, highlighting the transferability of our recipe.

Taken together, these results strongly suggest that \textbf{\ourmethod RL training teaches the model new capabilities and improved terminal use}, beyond overfitting on Terminal-Bench or our harness.

\section{Analysis \& Discussion}
\label{sec:discussion}

\paragraph{Qwen 3.5 does not benefit as much from existing SFT datasets.} 
\label{sec:sft_data_ablation}
Common post-training pipelines usually perform SFT training to `warm-start' the model before RL training~\citep{Guo_2025,olmo2026olmo3}. However, we find that existing datasets seem to degrade Qwen 3.5's performance by as much as 10 points, likely due to its already strong performance. In contrast, the same data can raise Qwen 3 8B's performance by as much as 9 points (see \S\ref{app:sft_qwen_ablations}). We leave further exploration of good SFT mixtures for Qwen 3.5 as future work.

\paragraph{Terminal agent RL training is difficult to stabilise.} Over the course of this work, we often found training unstable, with runs often collapsing past 300 steps.
We apply three changes to improve stability: (1) using an FP32 LM head, which reduced trainer-inference mismatch (\S\ref{app:lm_head}); (2) using larger group sizes, which alleviated collapse in initial setups (\S\ref{app:group_size}); (3) improving our sandboxing infrastructure. Surprisingly, we found that reducing noise from infrastructure could be the difference between collapses at 300 steps and smooth training to 1000 steps.

\begin{table}[t]
\centering
\small
\begin{tabular}{lcc}
\toprule
\textbf{Model} & \textbf{TB Lite} & \textbf{TB 2.1} \\
\midrule
Qwen 3.5 9B                 & $49.4 \pm 1.9$ & $20.5 \pm 1.7$ \\
\quad + RL (pre-repair) & $59.8 \pm 1.8$ & $27.3 \pm 1.4$ \\
\quad + RL (\ourdata)     & $\mathbf{65.6 \pm 2.0}$ & $\mathbf{29.6 \pm 1.8}$ \\
\quad + Qwen 3.5 9B-Generated RL (3.1k samples) & $62.0 \pm 2.0$ & $27.3 \pm 1.6$ \\
\bottomrule
\end{tabular}
\caption{\textbf{LM-judge repair improves downstream performance.} RL training of Qwen 3.5 9B on \ourdata before and after the LM-judge verification and repair pass (\S\ref{sec:gen_pipeline}), and using RL data generated by Qwen 3.5 9B itself (\S\ref{sec:discussion}).}
\label{tab:repair_ablation}
\end{table}

\paragraph{LM judge-based repair significantly improves data.} We compare training on our final dataset against the version before applying the final verification and repair step in Table~\ref{tab:repair_ablation}. We find that the repair step significantly improves the performance of models trained on it, likely due to reducing the likelihood that the model relies on shortcuts in the task (due to weak verifiers).

\paragraph{Models can self-improve using their own synthetic data.} We rerun our data generation pipeline (including classification and repair) entirely using Qwen 3.5 9B to construct a smaller dataset of 3.1k samples. In Table~\ref{tab:repair_ablation} we find that it still results in non-trivial improvements, even outperforming over half the baseline datasets compared in Table~\ref{tab:termbench}. This suggests that \textbf{our data pipeline and recipe do not require a strong teacher model, but can in fact be used with a model of similar strength to the policy model.} This may be due to information asymmetry: for many tasks, constructing the environment may be simpler than solving it (e.g., creating a sudoku solving task).

\section{Conclusion}

We present \ourmethod, a simple recipe for training terminal agents: \ourdata, a dataset of 13,565 RL environments generated with explicit control over difficulty and diversity, paired with an outcome-only RL recipe. \ourmodel achieves state-of-the-art Terminal-Bench 2.0 performance among open models under 10B parameters, and our recipe improves models across sizes, model families, harnesses, and out-of-domain tasks such as SWE-Bench Verified and AIME. We release all data, models, and code as a baseline for future open work on terminal agents.

\section*{Acknowledgments}

We thank members of UW NLP and the Open Ecosystem team at Ai2 for feedback and discussion throughout the project. We thank Michael Noukhovitch for useful discussions on RL stability and the Ai2 beaker team for help with infrastructure. We additionally thank Nebius for some cloud credits which helped with running evaluations as part of this work.







\bibliography{anthology-2,references}

@article{merrill2026terminal,
  title={Terminal-bench: Benchmarking agents on hard, realistic tasks in command line interfaces},
  author={Mike A. Merrill and Alexander G. Shaw and Nicholas Carlini and Boxuan Li and Harsh Raj and Ivan Bercovich and Lin Shi and Jeong Yeon Shin and Thomas Walshe and E. Kelly Buchanan and Junhong Shen and Guanghao Ye and Haowei Lin and Jason Poulos and Maoyu Wang and Marianna Nezhurina and Jenia Jitsev and Di Lu and Orfeas Menis Mastromichalakis and Zhiwei Xu and Zizhao Chen and Yue Liu and Robert Zhang and Leon Liangyu Chen and Anurag Kashyap and Jan-Lucas Uslu and Jeffrey Li and Jianbo Wu and Minghao Yan and Song Bian and Vedang Sharma and Ke Sun and Steven Dillmann and Akshay Anand and Andrew Lanpouthakoun and Bardia Koopah and Changran Hu and Etash Guha and Gabriel H. S. Dreiman and Jiacheng Zhu and Karl Krauth and Li Zhong and Niklas Muennighoff and Robert Amanfu and Shangyin Tan and Shreyas Pimpalgaonkar and Tushar Aggarwal and Xiangning Lin and Xin Lan and Xuandong Zhao and Yiqing Liang and Yuanli Wang and Zilong Wang and Changzhi Zhou and David Heineman and Hange Liu and Harsh Trivedi and John Yang and Junhong Lin and Manish Shetty and Michael Yang and Nabil Omi and Negin Raoof and Shanda Li and Terry Yue Zhuo and Wuwei Lin and Yiwei Dai and Yuxin Wang and Wenhao Chai and Shang Zhou and Dariush Wahdany and Ziyu She and Jiaming Hu and Zhikang Dong and Yuxuan Zhu and Sasha Cui and Ahson Saiyed and Arinbjörn Kolbeinsson and Jesse Hu and Christopher Michael Rytting and Ryan Marten and Yixin Wang and Alex Dimakis and Andy Konwinski and Ludwig Schmidt},
  journal={arXiv preprint arXiv:2601.11868},
  year={2026}
}

@misc{openthoughts-agent,
  author = {{OpenThoughts-Agent Team}},
  month = Dec,
  title = {{OpenThoughts-Agent}},
  howpublished = {https://www.open-thoughts.ai/blog/agent},
  year = {2025}
}

@misc{raoof2026openthoughtsagent,
      title={OpenThoughts-Agent: Data Recipes for Agentic Models}, 
      author={Negin Raoof and Richard Zhuang and Marianna Nezhurina and Etash Guha and Atula Tejaswi and Ryan Marten and Charlie F. Ruan and Tyler Griggs and Alexander Glenn Shaw and Hritik Bansal and E. Kelly Buchanan and Artem Gazizov and Reinhard Heckel and Chinmay Hegde and Sankalp Jajee and Daanish Khazi and Emmanouil Koukoumidis and Xiangyi Li and Hange Liu and Shlok Natarajan and Harsh Raj and Nicholas Roberts and Ethan Shen and Nishad Singhi and Michael Siu and Ashima Suvarna and Hanwen Xing and Patrick Yubeaton and Robert Zhang and Leon Liangyu Chen and Xiaokun Chen and Steven Dillmann and Saadia Gabriel and Xunyi Jiang and Anurag Kashyap and Boxuan Li and Yein Park and Minh Pham and Sujay Sanghavi and Lin Shi and Ke Sun and Yixin Wang and Zhiwei Xu and Erica Zhang and Siyan Zhao and Wanjia Zhao and Jenia Jitsev and Alex Dimakis and Benjamin Feuer and Ludwig Schmidt},
      year={2026},
      eprint={2606.24855},
      archivePrefix={arXiv},
      primaryClass={cs.AI},
      url={https://arxiv.org/abs/2606.24855}, 
}

@article{gandhi2025endless,
    title={Endless Terminals: Scaling RL Environments for Terminal Agents},
    author={Gandhi, Kanishk and Garg, Shivam and Goodman, Noah D. and Papailiopoulos, Dimitris},
    journal={arXiv preprint arXiv:2601.16443},
    year={2025}
}

@misc{qwen3.5,
    title  = {{Qwen3.5}: Towards Native Multimodal Agents},
    author = {{Qwen Team}},
    month  = {February},
    year   = {2026},
    url    = {https://qwen.ai/blog?id=qwen3.5}
}

@inproceedings{lin-etal-2018-nl2bash,
    title = "{NL}2{B}ash: A Corpus and Semantic Parser for Natural Language Interface to the Linux Operating System",
    author = "Lin, Xi Victoria  and
      Wang, Chenglong  and
      Zettlemoyer, Luke  and
      Ernst, Michael D.",
    editor = "Calzolari, Nicoletta  and
      Choukri, Khalid  and
      Cieri, Christopher  and
      Declerck, Thierry  and
      Goggi, Sara  and
      Hasida, Koiti  and
      Isahara, Hitoshi  and
      Maegaard, Bente  and
      Mariani, Joseph  and
      Mazo, H{\'e}l{\`e}ne  and
      Moreno, Asuncion  and
      Odijk, Jan  and
      Piperidis, Stelios  and
      Tokunaga, Takenobu",
    booktitle = "Proceedings of the Eleventh International Conference on Language Resources and Evaluation ({LREC} 2018)",
    month = may,
    year = "2018",
    address = "Miyazaki, Japan",
    publisher = "European Language Resources Association (ELRA)",
    url = "https://aclanthology.org/L18-1491/"
}

@misc{anthropic_claude_code,
  author       = {Anthropic},
  title        = {Claude Code: An Agentic Coding Tool},
  year         = {2025},
  howpublished = {\url{https://github.com/anthropics/claude-code}},
  note         = {Accessed: 2026-06-06}
}

@misc{research2026composer2technicalreport,
      title={Composer 2 Technical Report}, 
      author={{Cursor Research} and : and Aaron Chan and Ahmed Shalaby and Alexander Wettig and Aman Sanger and Andrew Zhai and Anurag Ajay and Ashvin Nair and Charlie Snell and Chen Lu and Chen Shen and Emily Jia and Federico Cassano and Hanpeng Liu and Haoyu Chen and Henry Wildermuth and Jacob Jackson and Janet Li and Jediah Katz and Jiajun Yao and Joey Hejna and Josh Warner and Julius Vering and Kevin Frans and Lee Danilek and Less Wright and Lujing Cen and Luke Melas-Kyriazi and Michael Truell and Michiel de Jong and Naman Jain and Nate Schmidt and Nathan Wang and Niklas Muennighoff and Oleg Rybkin and Paul Loh and Phillip Kravtsov and Rishabh Yadav and Sahil Shah and Sam Kottler and Alexander M Rush and Shengtong Zhang and Shomil Jain and Sriram Sankar and Stefan Heule and Stuart H. Sul and Sualeh Asif and Victor Rong and Wanqi Zhu and William Lin and Yuchen Wu and Yuri Volkov and Yury Zemlyanskiy and Zack Holbrook and Zhiyuan Zhang},
      year={2026},
      eprint={2603.24477},
      archivePrefix={arXiv},
      primaryClass={cs.SE},
      url={https://arxiv.org/abs/2603.24477}, 
}

@misc{cheng2026computerenvironmentselicitgeneral,
      title={Computer Environments Elicit General Agentic Intelligence in LLMs}, 
      author={Daixuan Cheng and Shaohan Huang and Yuxian Gu and Huatong Song and Guoxin Chen and Li Dong and Wayne Xin Zhao and Ji-Rong Wen and Furu Wei},
      year={2026},
      eprint={2601.16206},
      archivePrefix={arXiv},
      primaryClass={cs.CL},
      url={https://arxiv.org/abs/2601.16206}, 
}

@inproceedings{kwon2023efficient,
  title={Efficient Memory Management for Large Language Model Serving with PagedAttention},
  author={Woosuk Kwon and Zhuohan Li and Siyuan Zhuang and Ying Sheng and Lianmin Zheng and Cody Hao Yu and Joseph E. Gonzalez and Hao Zhang and Ion Stoica},
  booktitle={Proceedings of the ACM SIGOPS 29th Symposium on Operating Systems Principles},
  year={2023}
}

@misc{Harbor_Framework,
author = {{Harbor Framework Team}},
month = jan,
title = {{Harbor: A framework for evaluating and optimizing agents and models in container environments}},
url = {https://github.com/harbor-framework/harbor},
year = {2026}
}

@misc{minimax2025interleaved,
  title        = {Interleaved Thinking Unlocks Reliable {MiniMax-M2} Agentic Capability},
  author       = {{MiniMax}},
  year         = {2025},
  month        = nov,
  howpublished = {\url{https://www.minimax.io/news/why-is-interleaved-thinking-important-for-m2}},
  note         = {Blog post. Accessed June 2026}
}

@misc{yang2025swesmith,
  title={SWE-smith: Scaling Data for Software Engineering Agents}, 
  author={John Yang and Kilian Lieret and Carlos E. Jimenez and Alexander Wettig and Kabir Khandpur and Yanzhe Zhang and Binyuan Hui and Ofir Press and Ludwig Schmidt and Diyi Yang},
  year={2025},
  eprint={2504.21798},
  archivePrefix={arXiv},
  primaryClass={cs.SE},
  url={https://arxiv.org/abs/2504.21798},
}

@misc{nanorollout,
  title        = {{NanoRollout}: Scale digital agent rollouts without pain},
  author       = {Wang, Junli and Cheng, Zhoujun and Zhang, Yuxuan and Hao, Shibo and Tang, Yao and Hu, Zhiting and Ammanabrolu, Prithviraj and Zhang, Hao},
  year         = {2026},
  month        = may,
  howpublished = {\url{https://cocoa-org.notion.site/nanorollout}},
  note         = {Notion Blog}
}

@misc{OpenThoughts-TBLite,
  author = {{OpenThoughts-Agent Team}, Snorkel AI, Bespoke Labs},
  month = Feb,
  title = {{OpenThoughts-TBLite: A High-Signal Benchmark for Iterating on Terminal Agents}},
  howpublished = {https://www.openthoughts.ai/blog/openthoughts-tblite},
  year = {2026}
}

@misc{lin2026cligymscalableclitask,
      title={CLI-Gym: Scalable CLI Task Generation via Agentic Environment Inversion}, 
      author={Yusong Lin and Haiyang Wang and Shuzhe Wu and Lue Fan and Feiyang Pan and Sanyuan Zhao and Dandan Tu},
      year={2026},
      eprint={2602.10999},
      archivePrefix={arXiv},
      primaryClass={cs.AI},
      url={https://arxiv.org/abs/2602.10999}, 
}

@misc{minimax2025minimaxm1scalingtesttimecompute,
      title={MiniMax-M1: Scaling Test-Time Compute Efficiently with Lightning Attention}, 
      author={MiniMax and Aili Chen and Aonian Li and Bangwei Gong and Binyang Jiang and Bo Fei and Bo Yang and Boji Shan and Changqing Yu and Chao Wang and Cheng Zhu and Chengjun Xiao and Chengyu Du and Chi Zhang and Chu Qiao and Chunhao Zhang and Chunhui Du and Congchao Guo and Da Chen and Deming Ding and Dianjun Sun and Dong Li and Enwei Jiao and Haigang Zhou and Haimo Zhang and Han Ding and Haohai Sun and Haoyu Feng and Huaiguang Cai and Haichao Zhu and Jian Sun and Jiaqi Zhuang and Jiaren Cai and Jiayuan Song and Jin Zhu and Jingyang Li and Jinhao Tian and Jinli Liu and Junhao Xu and Junjie Yan and Junteng Liu and Junxian He and Kaiyi Feng and Ke Yang and Kecheng Xiao and Le Han and Leyang Wang and Lianfei Yu and Liheng Feng and Lin Li and Lin Zheng and Linge Du and Lingyu Yang and Lunbin Zeng and Minghui Yu and Mingliang Tao and Mingyuan Chi and Mozhi Zhang and Mujie Lin and Nan Hu and Nongyu Di and Peng Gao and Pengfei Li and Pengyu Zhao and Qibing Ren and Qidi Xu and Qile Li and Qin Wang and Rong Tian and Ruitao Leng and Shaoxiang Chen and Shaoyu Chen and Shengmin Shi and Shitong Weng and Shuchang Guan and Shuqi Yu and Sichen Li and Songquan Zhu and Tengfei Li and Tianchi Cai and Tianrun Liang and Weiyu Cheng and Weize Kong and Wenkai Li and Xiancai Chen and Xiangjun Song and Xiao Luo and Xiao Su and Xiaobo Li and Xiaodong Han and Xinzhu Hou and Xuan Lu and Xun Zou and Xuyang Shen and Yan Gong and Yan Ma and Yang Wang and Yiqi Shi and Yiran Zhong and Yonghong Duan and Yongxiang Fu and Yongyi Hu and Yu Gao and Yuanxiang Fan and Yufeng Yang and Yuhao Li and Yulin Hu and Yunan Huang and Yunji Li and Yunzhi Xu and Yuxin Mao and Yuxuan Shi and Yuze Wenren and Zehan Li and Zelin Li and Zhanxu Tian and Zhengmao Zhu and Zhenhua Fan and Zhenzhen Wu and Zhichao Xu and Zhihang Yu and Zhiheng Lyu and Zhuo Jiang and Zibo Gao and Zijia Wu and Zijian Song and Zijun Sun},
      year={2025},
      eprint={2506.13585},
      archivePrefix={arXiv},
      primaryClass={cs.CL},
      url={https://arxiv.org/abs/2506.13585}, 
}

@misc{olmo2026olmo3,
      title={Olmo 3}, 
      author={Team Olmo and : and Allyson Ettinger and Amanda Bertsch and Bailey Kuehl and David Graham and David Heineman and Dirk Groeneveld and Faeze Brahman and Finbarr Timbers and Hamish Ivison and Jacob Morrison and Jake Poznanski and Kyle Lo and Luca Soldaini and Matt Jordan and Mayee Chen and Michael Noukhovitch and Nathan Lambert and Pete Walsh and Pradeep Dasigi and Robert Berry and Saumya Malik and Saurabh Shah and Scott Geng and Shane Arora and Shashank Gupta and Taira Anderson and Teng Xiao and Tyler Murray and Tyler Romero and Victoria Graf and Akari Asai and Akshita Bhagia and Alexander Wettig and Alisa Liu and Aman Rangapur and Chloe Anastasiades and Costa Huang and Dustin Schwenk and Harsh Trivedi and Ian Magnusson and Jaron Lochner and Jiacheng Liu and Lester James V. Miranda and Maarten Sap and Malia Morgan and Michael Schmitz and Michal Guerquin and Michael Wilson and Regan Huff and Ronan Le Bras and Rui Xin and Rulin Shao and Sam Skjonsberg and Shannon Zejiang Shen and Shuyue Stella Li and Tucker Wilde and Valentina Pyatkin and Will Merrill and Yapei Chang and Yuling Gu and Zhiyuan Zeng and Ashish Sabharwal and Luke Zettlemoyer and Pang Wei Koh and Ali Farhadi and Noah A. Smith and Hannaneh Hajishirzi},
      year={2026},
      eprint={2512.13961},
      archivePrefix={arXiv},
      primaryClass={cs.CL},
      url={https://arxiv.org/abs/2512.13961}, 
}

@misc{yu2025dapoopensourcellmreinforcement,
      title={DAPO: An Open-Source LLM Reinforcement Learning System at Scale}, 
      author={Qiying Yu and Zheng Zhang and Ruofei Zhu and Yufeng Yuan and Xiaochen Zuo and Yu Yue and Weinan Dai and Tiantian Fan and Gaohong Liu and Lingjun Liu and Xin Liu and Haibin Lin and Zhiqi Lin and Bole Ma and Guangming Sheng and Yuxuan Tong and Chi Zhang and Mofan Zhang and Wang Zhang and Hang Zhu and Jinhua Zhu and Jiaze Chen and Jiangjie Chen and Chengyi Wang and Hongli Yu and Yuxuan Song and Xiangpeng Wei and Hao Zhou and Jingjing Liu and Wei-Ying Ma and Ya-Qin Zhang and Lin Yan and Mu Qiao and Yonghui Wu and Mingxuan Wang},
      year={2025},
      eprint={2503.14476},
      archivePrefix={arXiv},
      primaryClass={cs.LG},
      url={https://arxiv.org/abs/2503.14476}, 
}

@misc{shao2024deepseekmathpushinglimitsmathematical,
      title={DeepSeekMath: Pushing the Limits of Mathematical Reasoning in Open Language Models}, 
      author={Zhihong Shao and Peiyi Wang and Qihao Zhu and Runxin Xu and Junxiao Song and Xiao Bi and Haowei Zhang and Mingchuan Zhang and Y. K. Li and Y. Wu and Daya Guo},
      year={2024},
      eprint={2402.03300},
      archivePrefix={arXiv},
      primaryClass={cs.CL},
      url={https://arxiv.org/abs/2402.03300}, 
}

@misc{ge2025scalingsyntheticdatacreation,
      title={Scaling Synthetic Data Creation with 1,000,000,000 Personas}, 
      author={Tao Ge and Xin Chan and Xiaoyang Wang and Dian Yu and Haitao Mi and Dong Yu},
      year={2025},
      eprint={2406.20094},
      archivePrefix={arXiv},
      primaryClass={cs.CL},
      url={https://arxiv.org/abs/2406.20094}, 
}

@misc{qi2026rethinkingtrustregionllm,
      title={Rethinking the Trust Region in LLM Reinforcement Learning}, 
      author={Penghui Qi and Xiangxin Zhou and Zichen Liu and Tianyu Pang and Chao Du and Min Lin and Wee Sun Lee},
      year={2026},
      eprint={2602.04879},
      archivePrefix={arXiv},
      primaryClass={cs.LG},
      url={https://arxiv.org/abs/2602.04879}, 
}

@misc{lambert2025tulu3pushingfrontiers,
      title={Tulu 3: Pushing Frontiers in Open Language Model Post-Training}, 
      author={Nathan Lambert and Jacob Morrison and Valentina Pyatkin and Shengyi Huang and Hamish Ivison and Faeze Brahman and Lester James V. Miranda and Alisa Liu and Nouha Dziri and Shane Lyu and Yuling Gu and Saumya Malik and Victoria Graf and Jena D. Hwang and Jiangjiang Yang and Ronan Le Bras and Oyvind Tafjord and Chris Wilhelm and Luca Soldaini and Noah A. Smith and Yizhong Wang and Pradeep Dasigi and Hannaneh Hajishirzi},
      year={2025},
      eprint={2411.15124},
      archivePrefix={arXiv},
      primaryClass={cs.CL},
      url={https://arxiv.org/abs/2411.15124}, 
}

@article{Guo_2025,
   title={DeepSeek-R1 incentivizes reasoning in LLMs through reinforcement learning},
   volume={645},
   ISSN={1476-4687},
   url={http://dx.doi.org/10.1038/s41586-025-09422-z},
   DOI={10.1038/s41586-025-09422-z},
   number={8081},
   journal={Nature},
   publisher={Springer Science and Business Media LLC},
   author={Guo, Daya and Yang, Dejian and Zhang, Haowei and Song, Junxiao and Wang, Peiyi and Zhu, Qihao and Xu, Runxin and Zhang, Ruoyu and Ma, Shirong and Bi, Xiao and Zhang, Xiaokang and Yu, Xingkai and Wu, Yu and Wu, Z. F. and Gou, Zhibin and Shao, Zhihong and Li, Zhuoshu and Gao, Ziyi and Liu, Aixin and Xue, Bing and Wang, Bingxuan and Wu, Bochao and Feng, Bei and Lu, Chengda and Zhao, Chenggang and Deng, Chengqi and Ruan, Chong and Dai, Damai and Chen, Deli and Ji, Dongjie and Li, Erhang and Lin, Fangyun and Dai, Fucong and Luo, Fuli and Hao, Guangbo and Chen, Guanting and Li, Guowei and Zhang, H. and Xu, Hanwei and Ding, Honghui and Gao, Huazuo and Qu, Hui and Li, Hui and Guo, Jianzhong and Li, Jiashi and Chen, Jingchang and Yuan, Jingyang and Tu, Jinhao and Qiu, Junjie and Li, Junlong and Cai, J. L. and Ni, Jiaqi and Liang, Jian and Chen, Jin and Dong, Kai and Hu, Kai and You, Kaichao and Gao, Kaige and Guan, Kang and Huang, Kexin and Yu, Kuai and Wang, Lean and Zhang, Lecong and Zhao, Liang and Wang, Litong and Zhang, Liyue and Xu, Lei and Xia, Leyi and Zhang, Mingchuan and Zhang, Minghua and Tang, Minghui and Zhou, Mingxu and Li, Meng and Wang, Miaojun and Li, Mingming and Tian, Ning and Huang, Panpan and Zhang, Peng and Wang, Qiancheng and Chen, Qinyu and Du, Qiushi and Ge, Ruiqi and Zhang, Ruisong and Pan, Ruizhe and Wang, Runji and Chen, R. J. and Jin, R. L. and Chen, Ruyi and Lu, Shanghao and Zhou, Shangyan and Chen, Shanhuang and Ye, Shengfeng and Wang, Shiyu and Yu, Shuiping and Zhou, Shunfeng and Pan, Shuting and Li, S. S. and Zhou, Shuang and Wu, Shaoqing and Yun, Tao and Pei, Tian and Sun, Tianyu and Wang, T. and Zeng, Wangding and Liu, Wen and Liang, Wenfeng and Gao, Wenjun and Yu, Wenqin and Zhang, Wentao and Xiao, W. L. and An, Wei and Liu, Xiaodong and Wang, Xiaohan and Chen, Xiaokang and Nie, Xiaotao and Cheng, Xin and Liu, Xin and Xie, Xin and Liu, Xingchao and Yang, Xinyu and Li, Xinyuan and Su, Xuecheng and Lin, Xuheng and Li, X. Q. and Jin, Xiangyue and Shen, Xiaojin and Chen, Xiaosha and Sun, Xiaowen and Wang, Xiaoxiang and Song, Xinnan and Zhou, Xinyi and Wang, Xianzu and Shan, Xinxia and Li, Y. K. and Wang, Y. Q. and Wei, Y. X. and Zhang, Yang and Xu, Yanhong and Li, Yao and Zhao, Yao and Sun, Yaofeng and Wang, Yaohui and Yu, Yi and Zhang, Yichao and Shi, Yifan and Xiong, Yiliang and He, Ying and Piao, Yishi and Wang, Yisong and Tan, Yixuan and Ma, Yiyang and Liu, Yiyuan and Guo, Yongqiang and Ou, Yuan and Wang, Yuduan and Gong, Yue and Zou, Yuheng and He, Yujia and Xiong, Yunfan and Luo, Yuxiang and You, Yuxiang and Liu, Yuxuan and Zhou, Yuyang and Zhu, Y. X. and Huang, Yanping and Li, Yaohui and Zheng, Yi and Zhu, Yuchen and Ma, Yunxian and Tang, Ying and Zha, Yukun and Yan, Yuting and Ren, Z. Z. and Ren, Zehui and Sha, Zhangli and Fu, Zhe and Xu, Zhean and Xie, Zhenda and Zhang, Zhengyan and Hao, Zhewen and Ma, Zhicheng and Yan, Zhigang and Wu, Zhiyu and Gu, Zihui and Zhu, Zijia and Liu, Zijun and Li, Zilin and Xie, Ziwei and Song, Ziyang and Pan, Zizheng and Huang, Zhen and Xu, Zhipeng and Zhang, Zhongyu and Zhang, Zhen},
   year={2025},
   month=sep, pages={633–638} }

@misc{shen2026sera,
      title={SERA: Soft-Verified Efficient Repository Agents}, 
      author={Ethan Shen and Daniel Tormoen and Saurabh Shah and Ali Farhadi and Tim Dettmers},
      year={2026},
      eprint={2601.20789},
      archivePrefix={arXiv},
      primaryClass={cs.CL},
      url={https://arxiv.org/abs/2601.20789}, 
}

@misc{wang2026letflowagenticcrafting,
      title={Let It Flow: Agentic Crafting on Rock and Roll, Building the ROME Model within an Open Agentic Learning Ecosystem}, 
      author={Weixun Wang and XiaoXiao Xu and Wanhe An and Fangwen Dai and Wei Gao and Yancheng He and Ju Huang and Qiang Ji and Hanqi Jin and Xiaoyang Li and Yang Li and Zhongwen Li and Shirong Lin and Jiashun Liu and Zenan Liu and Tao Luo and Dilxat Muhtar and Yuanbin Qu and Jiaqiang Shi and Qinghui Sun and Yingshui Tan and Hao Tang and Runze Wang and Yi Wang and Zhaoguo Wang and Yanan Wu and Shaopan Xiong and Binchen Xu and Xander Xu and Yuchi Xu and Qipeng Zhang and Xixia Zhang and Haizhou Zhao and Jie Zhao and Shuaibing Zhao and Baihui Zheng and Jianhui Zheng and Suhang Zheng and Yanni Zhu and Mengze Cai and Kerui Cao and Xitong Chen and Yue Dai and Lifan Du and Tao Feng and Tao He and Jin Hu and Yijie Hu and Ziyu Jiang and Cheng Li and Xiang Li and Jing Liang and Xin Lin and Chonghuan Liu and ZhenDong Liu and Zhiqiang Lv and Haodong Mi and Yanhu Mo and Junjia Ni and Shixin Pei and Jingyu Shen and XiaoShuai Song and Cecilia Wang and Chaofan Wang and Kangyu Wang and Pei Wang and Tao Wang and Wei Wang and Ke Xiao and Mingyu Xu and Tiange Xu and Nan Ya and Siran Yang and Jianan Ye and Yaxing Zang and Duo Zhang and Junbo Zhang and Boren Zheng and Wanxi Deng and Ling Pan and Lin Qu and Wenbo Su and Jiamang Wang and Wei Wang and Hu Wei and Minggang Wu and Cheng Yu and Bing Zhao and Zhicheng Zheng and Bo Zheng},
      year={2026},
      eprint={2512.24873},
      archivePrefix={arXiv},
      primaryClass={cs.AI},
      url={https://arxiv.org/abs/2512.24873}, 
}

@misc{pan2024trainingsoftwareengineeringagents,
      title={Training Software Engineering Agents and Verifiers with SWE-Gym}, 
      author={Jiayi Pan and Xingyao Wang and Graham Neubig and Navdeep Jaitly and Heng Ji and Alane Suhr and Yizhe Zhang},
      year={2024},
      eprint={2412.21139},
      archivePrefix={arXiv},
      primaryClass={cs.SE},
      url={https://arxiv.org/abs/2412.21139}, 
}

@article{brown2020language,
  title={Language Models are Few-Shot Learners},
  author={Brown, Tom B. and Mann, Benjamin and Ryder, Nick and Subbiah, Melanie and Kaplan, Jared and Dhariwal, Prafulla and Neelakantan, Arvind and Shyam, Pranav and Sastry, Girish and Askell, Amanda and Agarwal, Sandhini and Herbert-Voss, Ariel and Krueger, Gretchen and Henighan, Tom and Child, Rewon and Ramesh, Aditya and Ziegler, Daniel M. and Wu, Jeffrey and Winter, Clemens and Hesse, Christopher and Chen, Mark and Sigler, Eric and Litwin, Mateusz and Gray, Scott and Chess, Benjamin and Clark, Jack and Berner, Christopher and McCandlish, Sam and Radford, Alec and Sutskever, Ilya and Amodei, Dario},
  journal={Advances in Neural Information Processing Systems},
  volume={33},
  pages={1877--1901},
  year={2020}
}

@misc{touvron2023llama,
      title={{LLaMA}: Open and Efficient Foundation Language Models},
      author={Hugo Touvron and Thibaut Lavril and Gautier Izacard and Xavier Martinet and Marie-Anne Lachaux and Timoth{\'e}e Lacroix and Baptiste Rozi{\`e}re and Naman Goyal and Eric Hambro and Faisal Azhar and Aurelien Rodriguez and Armand Joulin and Edouard Grave and Guillaume Lample},
      year={2023},
      eprint={2302.13971},
      archivePrefix={arXiv},
      primaryClass={cs.CL},
      url={https://arxiv.org/abs/2302.13971}
}

@misc{badertdinov2025swerebenchautomatedpipelinetask,
      title={SWE-rebench: An Automated Pipeline for Task Collection and Decontaminated Evaluation of Software Engineering Agents}, 
      author={Ibragim Badertdinov and Alexander Golubev and Maksim Nekrashevich and Anton Shevtsov and Simon Karasik and Andrei Andriushchenko and Maria Trofimova and Daria Litvintseva and Boris Yangel},
      year={2025},
      eprint={2505.20411},
      archivePrefix={arXiv},
      primaryClass={cs.SE},
      url={https://arxiv.org/abs/2505.20411}
}

@article{pi2026data,
  title={On Data Engineering for Scaling LLM Terminal Capabilities},
  author={Pi, Renjie and Lam, Grace and Shoeybi, Mohammad and Jannaty, Pooya and Catanzaro, Bryan and Ping, Wei},
  journal={arXiv preprint arXiv:2602.21193},
  year={2026}
}

@article{zhu2026termigen,
  title={TermiGen: High-Fidelity Environment and Robust Trajectory Synthesis for Terminal Agents},
  author={Zhu, Kaijie and Nie, Yuzhou and Li, Yijiang and Huang, Yiming and Wu, Jialian and Liu, Jiang and Sun, Ximeng and Yin, Zhenfei and Wang, Lun and Liu, Zicheng and others},
  journal={arXiv preprint arXiv:2602.07274},
  year={2026}
}

@article{wu2026termitraj,
  title={Large-Scale Terminal Agentic Trajectory Generation from Dockerized Environments},
  author={Wu, Siwei and Li, Yizhi and Song, Yuyang and Zhang, Wei and Wang, Yang and Batista-Navarro, Riza and Yang, Xian and Tang, Mingjie and Dai, Bryan and Yang, Jian and others},
  journal={arXiv preprint arXiv:2602.01244},
  year={2026}
}

@article{pichai2025new,
  title={A new era of intelligence with Gemini 3},
  author={Pichai, Sundar and Hassabis, Demis and Kavukcuoglu, Koray},
  journal={{Mountain View, CA: Google. Available online at: https://blog.google/products-and-platforms/products/gemini/gemini-3 (Accessed February 1, 2026)}},
  year={2025}
}

@misc{chen2026sweuniversescalerealworldverifiable,
      title={SWE-Universe: Scale Real-World Verifiable Environments to Millions}, 
      author={Mouxiang Chen and Lei Zhang and Yunlong Feng and Xuwu Wang and Wenting Zhao and Ruisheng Cao and Jiaxi Yang and Jiawei Chen and Mingze Li and Zeyao Ma and Hao Ge and Zongmeng Zhang and Zeyu Cui and Dayiheng Liu and Jingren Zhou and Jianling Sun and Junyang Lin and Binyuan Hui and Zhongxin Liu},
      year={2026},
      eprint={2602.02361},
      archivePrefix={arXiv},
      primaryClass={cs.SE},
      url={https://arxiv.org/abs/2602.02361}, 
}

@inproceedings{
    jimenez2024swebench,
    title={{SWE}-bench: Can Language Models Resolve Real-world Github Issues?},
    author={Carlos E Jimenez and John Yang and Alexander Wettig and Shunyu Yao and Kexin Pei and Ofir Press and Karthik R Narasimhan},
    booktitle={The Twelfth International Conference on Learning Representations},
    year={2024},
    url={https://openreview.net/forum?id=VTF8yNQM66}
}

@article{team2025kimi,
  title={Kimi k1. 5: Scaling reinforcement learning with llms},
  author={{Kimi Team} and Du, Angang and Gao, Bofei and Xing, Bowei and Jiang, Changjiu and Chen, Cheng and Li, Cheng and Xiao, Chenjun and Du, Chenzhuang and Liao, Chonghua and others},
  journal={arXiv preprint arXiv:2501.12599},
  year={2025}
}

@article{yang2024swe,
  title={Swe-agent: Agent-computer interfaces enable automated software engineering},
  author={Yang, John and Jimenez, Carlos E and Wettig, Alexander and Lieret, Kilian and Yao, Shunyu and Narasimhan, Karthik and Press, Ofir},
  journal={Advances in Neural Information Processing Systems},
  volume={37},
  pages={50528--50652},
  year={2024}
}

@misc{yang2026terminallego,
  title={What Makes Interaction Trajectories Effective for Training Terminal Agents?},
  author={Sidi Yang and Chaofan Tao and Jierun Chen and Tiezheng Yu and Ruoyu Wang and Yuxin Jiang and Yiming Du and Wendong Xu and Jing Xiong and Taiqiang Wu and Lifeng Shang and Xiaohui Li and Ngai Wong and Haoli Bai},
  year={2026},
  eprint={2606.03461},
  archivePrefix={arXiv},
  primaryClass={cs.AI},
  url={https://arxiv.org/abs/2606.03461}
}

@misc{li2026recursivesynthesislonghorizonterminal,
      title={Recursive Synthesis for Long-Horizon Terminal Tasks}, 
      author={Zhongzhi Li and Yucheng Shi and Zongxia Li and Ruhan Wang and Anhao Li and Zixun Huang and Junyao Yang and Lei Ke and Ninghao Liu and Haitao Mi and Leowei Liang},
      year={2026},
      eprint={2608.05466},
      archivePrefix={arXiv},
      primaryClass={cs.AI},
      url={https://arxiv.org/abs/2608.05466}, 
}

@misc{yao2026river,
  title  = {Learning Generalizable Behaviors for Terminal Agents},
  author = {Yao, Yihang and Pang, Bo and Nguyen, Xuan Phi and Zhao, Ding and Joty, Shafiq and Yavuz, Semih},
  year   = {2026},
  note   = {arXiv preprint arXiv:2608.22631}
}

@misc{wu2026terminaluniverseturningagenttrajectories,
      title={Terminal-Universe: Turning Agent Trajectories into Scalable Terminal Environments}, 
      author={Jie Wu and Zhenru Zhang and Beichen Zhang and Xuwu Wang and Yuhui Su and Mouxiang Chen and Peng Wang and Zhihai Wang and Que Shen and Hao Zhou and An Yang and Fei Huang and Yujiu Yang and Dayiheng Liu},
      year={2026},
      eprint={2609.04148},
      archivePrefix={arXiv},
      primaryClass={cs.AI},
      url={https://arxiv.org/abs/2609.04148}, 
}

@article{peng2026litecoderterminal,
  title={LiteCoder-Terminal: Scaling Long-Horizon Terminal Environments for Learning Language Agents},
  author={Peng, Xiaoxuan and Zhang, Kaiqi and Lu, Xinyu and Cao, Boxi and Lu, Yaojie and Lin, Hongyu and Han, Xianpei and Sun, Le},
  journal={arXiv preprint arXiv:2605.29559},
  year={2026}
}

@misc{shi2026facet,
  title   = {{FACET}: Preserving Source Intent and Executable
             State in Terminal Task Synthesis},
  author  = {Kou Shi and Zun Wang and Qisheng Su and
             Shiting Huang and Ziao Zhang and Zhen Fang and
             Qingnan Ren and Jin Liu and Yu Zeng and
             Yiming Zhao and Lin Chen and Zehui Chen and Feng Zhao},
  year    = {2026},
  eprint  = {2608.18580},
  archivePrefix = {arXiv},
  primaryClass  = {cs.AI},
  url     = {https://arxiv.org/abs/2608.18580}
}

@misc{shen2026seta,
  author        = {Qijia Shen and Zhiqi Huang and Vamsidhar Kamanuru and Aznaur Aliev and Jay Rainton and Ahmed Awelkair and Zhichen Zeng and Jiajun Li and Shi Dong and Yueming Yuan and Boyuan Ma and Qizheng Zhang and Jiwei Fu and Yuzhen Mao and Wendong Fan and Ping Nie and Philip Torr and Bernard Ghanem and Changran Hu and Jonathan Lingjie Li and Urmish Thakker and Guohao Li},
  title         = {{SETA: Scaling Environments for Terminal Agents}},
  year          = {2026},
  month         = jul,
  eprint        = {2607.10891},
  archivePrefix = {arXiv},
  primaryClass  = {cs.AI},
  url           = {https://arxiv.org/abs/2607.10891}
}

@misc{meng2026calibforge,
  title         = {CalibForge: Adversarial Solver Calibration for Scaling Learnable Terminal Tasks},
  author        = {Fanzhe Meng and Guoxin Chen and Jiale Zhao and Shuang Sun and Zhiyu Lin and Wayne Xin Zhao and Ruihua Song and Ji-Rong Wen and Kai Jia},
  year          = {2026},
  eprint        = {2608.06352},
  archivePrefix = {arXiv},
  primaryClass  = {cs.LG},
  url           = {https://arxiv.org/abs/2608.06352}
}
\bibliographystyle{iclr2027_conference}

\appendix
\newpage
\appendix

\section{Limitations}
Our dataset generation pipeline is completely synthetic, and relies on the presence of a strong generator model. 
While we find positive results using Qwen 3.5 9B both as generator and policy, it is unclear if even stronger models or weaker models would behave similarly.
We also believe that our approach likely struggles on tasks involving large-scale codebases, as these are inherently difficult to generate (but well-covered by SWE-bench-style datasets).
Additionally, while we have focused on training smaller terminal agents as a baseline for academic settings, we note that running many isolated containers still proves expensive and/or difficult in open frameworks at scale, limiting training speed and efficiency and potentially still putting terminal agent training out of reach for academic groups.
Finally, our results may be lower than possible SOTA due to the shorter context length and simple harness used relative to strong industry approaches, although we see this as a feature of our approach that makes it more friendly to smaller teams and easier to develop with.

\section{Related Work}
\label{app:related_work}

\subsection{Data for Terminal Agents}
Early work on terminal agents focused on training models to translate natural language to bash statements. NL2Bash~\citep{lin-etal-2018-nl2bash} paired short instructions with bash commands mined from the web, and was converted into an RL dataset for the training of OpenThinker Agent~\citep{openthoughts-agent}. However, it failed to yield significant improvements over their strong SFT model. This showcases the need for generating more complex terminal-agent tasks, which provide stronger learning signals on more complex tasks. More recently, \citet{raoof2026openthoughtsagent} compared multiple RL datasets and found that the choice of RL dataset could have a significant impact on final performance, motivating more work into improving data for RL training specifically.

Recent work constructs terminal training data from various sources. Repository-based approaches include SWE-Smith~\citep{yang2025swesmith}, which synthesises bug-fixing instances, and TerminalTraj~\citep{wu2026termitraj}, which builds Dockerised tasks and verified trajectories from repositories. Related software-engineering datasets include SWE-Universe~\citep{chen2026sweuniversescalerealworldverifiable}, SWE-rebench~\citep{badertdinov2025swerebenchautomatedpipelinetask}, SWE-Gym~\citep{pan2024trainingsoftwareengineeringagents}, and SERA~\citep{shen2026sera}. CLI-Gym~\citep{lin2026cligymscalableclitask} extends repository-based task creation by perturbing agent paths through repository-based environments to construct new environments and tasks. Terminal-Lego~\citep{yang2026terminallego} and SETA-Env~\citep{shen2026seta} scrape sites such as StackOverflow and Stack Exchange for questions with verified answers and turn them into environments.
SETA additionally applies an evolutionary procedure to control diversity and difficulty.
FACET~\citep{shi2026facet} constructs environments and tasks from skill files, generating task instructions and verifiers after environment construction to ensure validity. 
Terminal-Universe~\citep{wu2026terminaluniverseturningagenttrajectories} constructs environments from public agent trajectories, but does not release any data.
We find in many cases that repository-based task creation results in unbalanced task splits in the data: for example, in Figure~\ref{fig:composition} both TerminalTraj and CLI-Gym are mostly comprised of software engineering tasks.

Another set of recent work generates data from a given taxonomy or set of seed tasks, and is the approach we take in this work.
We are particularly inspired by Endless Terminals~\citep{gandhi2025endless}, which
similarly utilises a strong external model to generate new terminal tasks, but produces significantly fewer tasks that are largely focused on file manipulation, making them too easy for modern models. For instance, we find that Gemini-3-Flash achieves over 90\% pass@1 on Endless Terminals.
TermiGen~\citep{zhu2026termigen} and LiteCoder-Terminal~\citep{peng2026litecoderterminal} similarly synthesise containers and tasks from seed prompts, using a fairly complex multi-agent pipeline to generate and validate the tasks, and do not explicitly control the difficulty of their tasks.
CalibForge~\citep{meng2026calibforge} instructs an agent to construct a task from browsing the web, and then iteratively improves the sandbox until it is on the edge of solvability.
Similarly, Recursive Synthetic Terminal Tasks (RST)~\citep{li2026recursivesynthesislonghorizonterminal} starts from a set of seed tasks, and iteratively rewrites them based on sandbox feedback to improve their difficulty and quality.
Nemotron-Terminal~\citep{pi2026data} both generates new data from seed prompts and adapts existing datasets into terminal tasks, but does not release any RL environment data, and similarly only validates their data through large-scale SFT training.
We find that these other datasets do not appear to improve over \ourdata, despite its much simpler generation pipeline, as shown in Table~\ref{tab:termbench}.

\section{Terminal Data Generation Pipeline}
\begin{table*}[t]
    \centering
    \caption{Composition axes used by our task-generation pipeline. The first three
    are seeded from the skill taxonomy of \citet{pi2026data}; the remaining six
    are our contributions, targeting diversity and difficulty.}
    \label{tab:axes}
    \begin{minipage}{0.97\textwidth}
    \renewcommand{\arraystretch}{1.4}
    \begin{tabular}{lll>{\raggedright\arraybackslash}p{0.32\textwidth}}
    \toprule
    \textbf{Axis} & \textbf{Source} & \textbf{Cardinality} & \textbf{Values} \\
    \midrule
    Domain               & \citet{pi2026data} & 9                 & security, software\_engineering, file\_operations, data\_querying, data\_science, debugging, scientific\_computing, data\_processing, system\_administration \\[4pt]
    Skill type           & \citet{pi2026data} & 4--7 / domain     & e.g.\ Algorithmic, Systems, Data Processing, Web Security, Testing, Mathematical, Multi-Language \\[4pt]
    Primitive skills     & \citet{pi2026data} & 20--40 / domain   & 3--5 primitive skills sampled per task (e.g.\ ``graph traversal and dependency resolution'', ``cryptographic hashing and checksum verification'') \\
    \addlinespace[8pt]
    Persona              & Ours               & 6--18 / domain    & domain-tied user roles (e.g.\ ``incident responder investigating a 3am page'', ``bioinformatics analyst processing sequences'') \\[4pt]
    Language             & Ours               & 8 (weighted)      & Python, C, Bash, C++, Rust, Go, multi-language, any \\[4pt]
    Task complexity      & Ours               & 4                 & short, moderate, complex, intricate (30--60 commands) \\[4pt]
    Command complexity   & Ours               & 3                 & bash-only; bash + code; bash + code + system services \\[4pt]
    Fixture              & Ours               & 7                 & text\_only, image, audio, video, stripped\_binary, vendored\_package, multi\_service\_compose \\[4pt]
    Verifier             & Ours               & 5                 & exact\_text, metric\_threshold, adversarial\_corpus, fuzz\_equivalence, multi\_protocol \\
    \bottomrule
    \end{tabular}
    \end{minipage}
\end{table*}

\subsection{Verifier and fixture kinds}
Each task independently samples one verifier kind and one fixture kind (Table~\ref{tab:axes}). The legacy defaults (\texttt{exact\_text}, \texttt{text\_only}) reproduce standard text-in / text-out RL tasks, while the remaining kinds add graded verification (Table~\ref{tab:verifier-kinds}) and non-text inputs (Table~\ref{tab:fixture-kinds}). Graded verifiers relax brittle string equality and expose a continuous difficulty knob, and non-text fixtures broaden task inputs while keeping the policy text-only---the agent processes each artifact through standard terminal tooling rather than native perception.

\begin{table*}[t]
    \centering
    \caption{Verifier kinds. Beyond legacy exact-text equality, we add four graded
    verifiers that relax exact matching and expose an explicit difficulty knob.}
    \label{tab:verifier-kinds}
    \renewcommand{\arraystretch}{1.3}
    \resizebox{\columnwidth}{!}{%
    \begin{tabular}{l>{\raggedright\arraybackslash}p{0.34\textwidth}>{\raggedright\arraybackslash}p{0.22\textwidth}>{\raggedright\arraybackslash}p{0.20\textwidth}}
    \toprule
    \textbf{Kind} & \textbf{Pass criterion} & \textbf{Example} & \textbf{Difficulty knob} \\
    \midrule
    \texttt{exact\_text}         & Output exactly equals the ground-truth string. & file contents match a reference & --- (legacy) \\
    \texttt{metric\_threshold}   & A numeric metric against a reference meets a threshold. & image SSIM $\geq 0.95$; speedup $\geq 1.3\times$ & threshold value \\
    \texttt{adversarial\_corpus} & Solution rejects every item in a malicious corpus \emph{and} preserves every item in a benign one. & a sanitiser that blocks exploits yet leaves clean inputs intact & per-corpus pass rate; corpus size \\
    \texttt{fuzz\_equivalence}   & Agent program matches a reference oracle bit-exactly on $N$ random inputs. & reproduce a stripped binary's output & $N$; input distribution \\
    \texttt{multi\_protocol}     & A service the agent brings up answers real protocol-level requests correctly. & HTTP / TCP / gRPC / SMTP request--response checks & number of protocols / conditions \\
    \bottomrule
    \end{tabular}}
\end{table*}

\begin{table*}[t]
    \centering
    \caption{Fixture kinds. Each task optionally ships a non-text artifact; the agent
    recovers the hidden ground truth via standard terminal tooling, so the policy
    stays text-only.}
    \label{tab:fixture-kinds}
    \renewcommand{\arraystretch}{1.3}
    \resizebox{\columnwidth}{!}{%
    \begin{tabular}{l>{\raggedright\arraybackslash}p{0.22\textwidth}>{\raggedright\arraybackslash}p{0.26\textwidth}>{\raggedright\arraybackslash}p{0.24\textwidth}}
    \toprule
    \textbf{Kind} & \textbf{Artifact} & \textbf{Agent tooling} & \textbf{Recovers} \\
    \midrule
    \texttt{text\_only}             & none & --- & --- (legacy) \\
    \texttt{image}                  & PNG / JPEG & OCR (tesseract), vision & hidden text / structure \\
    \texttt{audio}                  & WAV / MP3 & transcription (whisper.cpp, ffmpeg) & transcript / events \\
    \texttt{video}                  & MP4 & frame extraction (ffmpeg) + analysis & frame ranges, counts, detections \\
    \texttt{stripped\_binary}       & stripped / UPX-packed binary & objdump, gdb, strings; or black-box oracle & implemented algorithm \\
    \texttt{vendored\_package}      & pre-vendored package source with a perturbation & build / debug tools (no internet) & known-good code path after fix \\
    \texttt{multi\_service\_compose} & multiple cooperating services & service config / glue & end-to-end protocol flow \\
    \bottomrule
    \end{tabular}}
\end{table*}

\subsection{Rubric-based Data Repair}
\label{app:repair_stats}

We follow \citet{yao2026river} in classifying \ourdata using an LM judge by the following labels:
\begin{itemize}
    \item \texttt{VERIFIER-TOO-WEAK}: The verifier does not fully test the final state, allowing the model to pass without fully completing the task.
    \item \texttt{ANSWER-LEAK}: The answer or tests are in the initial image state, allowing trivial task completion.
    \item \texttt{CLEAN}: No issues detected.
    \item \texttt{INSTR-VERIFIER-MISMATCH}: The instruction and verifier contradict each other.
    \item \texttt{INSTR-ENV-MISMATCH}: The instruction and environment contradict each other.
    \item \texttt{OTHER}: Some other issue not captured by the reasons above.
    \item \texttt{INSTR-AMBIGUOUS}: The instruction is ambiguous.
    \item \texttt{TASK-TRIVIAL}: The task is trivial to perform.
\end{itemize}

We classify our initial 14.6K samples using a GPT 5.6 Sol judge, resulting in the classification seen in Table~\ref{tab:repair_rubric_labels}. We then prompt GPT 5.6 Sol to fix the top three issues (verifier too weak, answer leakage, and instruction-verifier mismatch) by producing either new, hardened verifiers or cleaning the image specification. We drop all other problematic tasks from the dataset.

\begin{table}[H]
\centering
\small
\begin{tabular}{@{}lrrr@{}}
\toprule
\textbf{label} & \textbf{Tasks} & \textbf{Share (\%)} & \textbf{Applied repairs} \\
\midrule
\texttt{VERIFIER-TOO-WEAK} & 11,160 & 76.4 & 10,990 \\
\texttt{ANSWER-LEAK} & 1,592 & 10.9 & 1,018 \\
\texttt{CLEAN} & 1,125 & 7.7 & - \\
\texttt{INSTR-VERIFIER-MISMATCH} & 441 & 3.0 & 432 \\
\texttt{INSTR-ENV-MISMATCH} & 177 & 1.2 & 0 \\
\texttt{OTHER} & 83 & 0.6 & 0 \\
\texttt{INSTR-AMBIGUOUS} & 23 & 0.2 & 0 \\
\texttt{TASK-TRIVIAL} & 0 & 0.0 & 0 \\
\midrule
\textbf{Total} & \textbf{14,601} & \textbf{100.0} & \textbf{12,440} \\
\bottomrule
\end{tabular}
\caption{GPT 5.6-judged rubric labels from the original 14,601 tasks and applied repair counts. We retain 12,440 tasks with an applied repair plus 1,125 originally \texttt{CLEAN} tasks without a patch, giving 13,565 tasks in the hardened prefilter dataset; the other 1,036 are excluded. Applied repairs do not imply a second rubric classification.}
\label{tab:repair_rubric_labels}
\end{table}

\subsection{Pass@k difficulty curves}
\label{app:pass_k_difficulty_curves}
\begin{figure}[t]
    \centering
    \includegraphics[width=.7\linewidth]{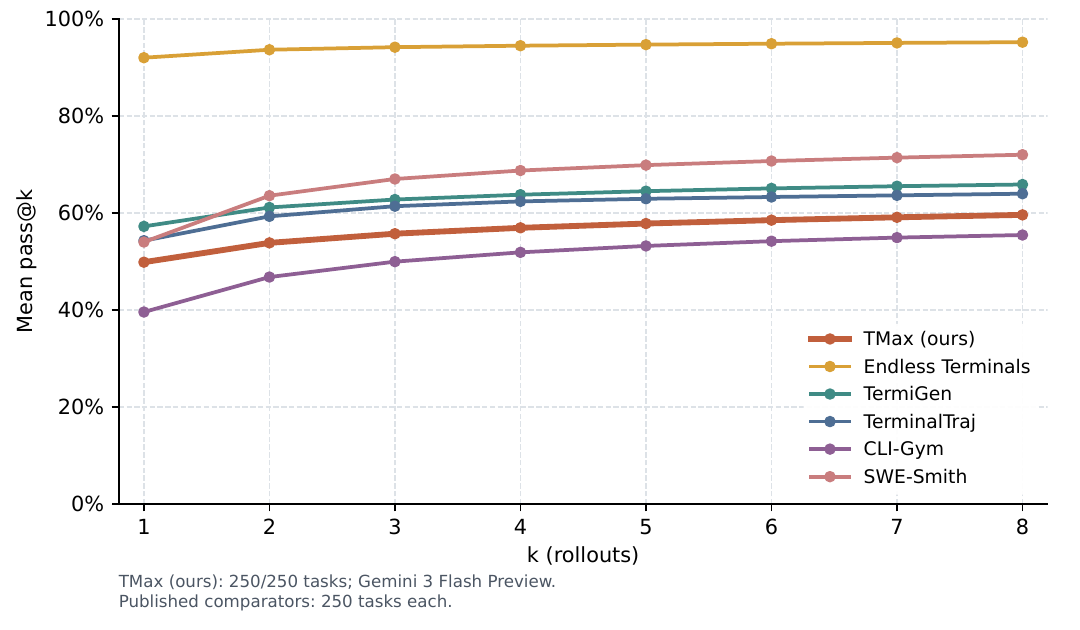}
    \caption{\textbf{Pass@$k$ difficulty curves.} Mean pass@$k$ ($k=1$--$8$) for Gemini-3-Flash-Preview on a fixed 250-task subsample per dataset (8 rollouts each); lower is harder. \ourmethod{} occupies the hardest band together with CLI-Gym and attains the second lowest pass@8 of any dataset, showing that its difficulty persists as $k$ grows, whereas easy datasets such as Endless-Terminals saturate near the ceiling.}
    \label{fig:pass-k}
\end{figure}

Table~\ref{tab:terminal-data-difficulty} reports pass@$k$ at $k\in\{1,4,8\}$; Figure~\ref{fig:pass-k} plots the full curve over $k=1$--$8$ for every dataset. The vertical position of a curve measures absolute difficulty, while its slope measures how much additional sampling recovers. \ourmethod{} occupies the hardest band together with CLI-Gym across all $k$, so its tasks remain unsolved even with eight rollouts, suggesting high difficulty and low rollout variance (although some tasks may still be infeasible).
Easy datasets behave very differently: Endless-Terminals already sits near the ceiling at $k=1$ and saturates almost immediately, leaving little headroom for a model to improve. This persistent, well-separated difficulty is what makes our data a useful source of learning signal for RL.

\subsection{Balance score}
\label{app:balance}
The balance score in \eqref{eq:balance} is the normalized effective number of categories of an empirical distribution. The numerator $\exp(H)$ is the size a uniform distribution would need to have the same Shannon entropy as the observed mass vector; dividing by the bucket count $N$ rescales the result to $[1/N, 1]$. A value of $1.0$ thus corresponds to perfectly uniform coverage of all buckets, while $1/N$ corresponds to all mass concentrated on a single bucket. We prefer this to raw entropy because it is comparable across axes with different bucket counts and reads as a natural ``fraction-of-uniform'' diversity score.

\subsection{Decontamination}
\label{app:decontamination}
We quantify potential train--test contamination as $n$-gram overlap between each dataset's task descriptions and the task descriptions of our evaluation benchmarks, Terminal-Bench~2.0 and TB-Lite. Using a sliding window of $n{=}13$ tokens (stride~1), we flag a dataset task as overlapping if any of its windows matches at least one $13$-gram from a benchmark task, following standard contamination protocols~\citep{brown2020language, touvron2023llama}; larger $n$ yields a stricter test with fewer spurious matches. Table~\ref{tab:terminal-data-decontamination} reports the fraction of each dataset flagged this way. Overlap is negligible across the board: \ourmethod{} shows $0\%$ overlap with both benchmarks, and only TerminalTraj exhibits any measurable overlap ($0.2\%$ on TB-Lite and $0.5\%$ on TB2). Since Terminal-Bench 2.1 is a modified version of 2.0, we do not run additional decontamination against it.

\begin{table*}[t]
\centering
\vspace{0.2em}
\resizebox{0.95\textwidth}{!}{%
\begin{tabular}{@{\hspace{0.4em}}l r r r r r r r@{\hspace{0.4em}}}
\toprule
\textbf{Benchmark} &
\shortstack{\textbf{TMax}\\\textbf{(ours)}} &
\shortstack{\textbf{Endless}\\\textbf{Terminals}} &
\shortstack{\textbf{OpenThinker}\\\textbf{Agent}} &
\shortstack{\textbf{TermiGen}} &
\shortstack{\textbf{Terminal}\\\textbf{Traj}} &
\shortstack{\textbf{CLI-Gym}} &
\shortstack{\textbf{SWE-Smith}} \\
\midrule
TB-Lite (openthoughts-tblite@2.0) & 0.0\% & 0.0\% & 0.0\% & 0.0\% & 0.2\% & 0.0\% & 0.0\% \\
TB2 (terminal-bench@2.0)          & 0.0\% & 0.0\% & 0.0\% & 0.0\% & 0.5\% & 0.0\% & 0.0\% \\
\bottomrule
\end{tabular}%
}
\caption{\textbf{Decontamination via $n$-gram overlap on task descriptions} ($n{=}13$, stride${=}1$). Each cell is the percentage of a dataset's task descriptions whose $n$-gram window overlaps at least one $n$-gram from the benchmark's task descriptions. Larger $n$ is stricter (fewer false positives).}
\label{tab:terminal-data-decontamination}
\end{table*}
\vspace{-0.3em}

\section{Harness Choices}
\label{app:harness_choices}

\begin{table}[t]
\centering
\begin{tabular}{llc}
\toprule
Model & Harness & TB Lite \\
\midrule
Haiku 4.5 & Ours   & $60.8 \pm 0.6$ \\
Haiku 4.5 & mini-SWE-agent         & \boldmath{$61.4 \pm 1.0$} \\
Haiku 4.5 & Terminus-2             & $56.1 \pm 3.9$ \\
\bottomrule
\end{tabular}
\caption{Claude Haiku 4.5 Terminal-Bench Lite performance when using different harnesses. Our harness is an earlier version from during development.}
\label{tab:haiku_tblite}
\end{table}

When deciding on an initial harness, we opted for a mini-SWE-agent inspired harness based on results over smaller closed-source offerings such as Claude Haiku 4.5. In Table~\ref{tab:haiku_tblite}, we show that our harness and mini-SWE-agent outperform Terminus-2, the usual default for Terminal-Bench models. We attribute this drop to more complex tool formats required for Terminus-2, which are especially difficult for smaller models to follow. Additionally, we wished to opt for a simpler harness to reduce complexity during RL training.

\section{Additional RL Training Details}

\subsection{Full RL training hyperparameters}
\label{app:full_rl_hypers}
\label{app:group_size}
\begin{table*}[t]
\centering
\small
\begin{tabular}{lll}
\toprule
\textbf{Category} & \textbf{Hyperparameter} & \textbf{Value} \\
\midrule
Model & Model dtype & \texttt{bfloat16} \\
Model & Gradient checkpointing & true \\
Model & LM head fp32 & true \\
\midrule
Data & Max prompt tokens & 2048 \\
Data & Per-turn max tokens & 16384 \\
Data & Max total response length & 65536 (32768 for Qwen 3 8B) \\
\midrule
Rollout & Unique prompts per rollout batch & 8 \\
Rollout & Group size & 32 \\
Rollout & Max async steps & 4 \\
Rollout & Active sampling & true \\
Rollout & Filter zero-std samples & true \\
Rollout & Sampling Temperature & 1.0 \\
\midrule
Optimisation & Optimiser & AdamW \\
Optimisation & Training steps & 500 \\
Optimisation & Learning rate & $1\times 10^{-6}$ \\
Optimisation & LR scheduler & constant \\
Optimisation & Max grad norm & 1.0 \\
\midrule
RL objective & Loss & DPPO \\
RL objective & Advantage normalisation & centered \\
RL objective & KL coefficient $\beta$ & 0.0 \\
RL objective & DPPO divergence & binary TV \\
RL objective & DPPO divergence threshold & 0.1 \\
\midrule
Tools & Max tool/env steps & 64 \\
Tools & Bash tool timeout & 120 s \\
\bottomrule
\end{tabular}
\caption{Hyperparameters for RL runs. We use these hyperparameters unless otherwise stated.}
\label{tab:full_rl_hypers}
\end{table*}

\begin{figure}[t]
    \centering
    \includegraphics[width=0.6\linewidth]{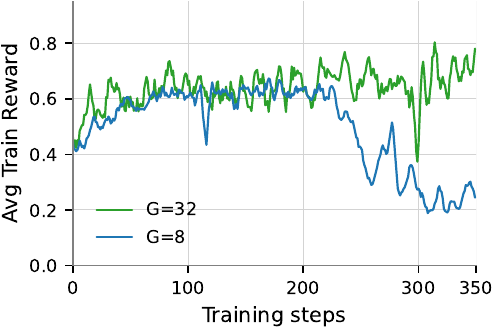}
    \caption{\textbf{Larger group sizes limit training collapse.} Average training reward when doing RL training on \ourdata{} with DPPO in initial experiments with a slightly different harness and Podman-based backend, using group size 8 vs.\ 32 (using 32 or 8 prompts per batch to keep total batch size constant).}
    \label{fig:group_size}
\end{figure}

Unless otherwise stated, we use the hyperparameters for RL training stated in Table~\ref{tab:full_rl_hypers}. We found that using a larger group size aided with stability in earlier experiments, as shown in Figure~\ref{fig:group_size}. We additionally found in early pilot experiments that DPPO demonstrated superior stability over GRPO, motivating us to stick with it for ablations.

\subsection{Full SFT Training Details}
\label{app:full_sft_hypers}
\begin{table*}[t]
\centering
\small
\begin{tabular}{lll}
\toprule
\textbf{Category} & \textbf{Hyperparameter} & \textbf{Value} \\
\midrule
Model & Precision & bf16 \\
Model & Attention implementation & flash attention 3 \\
Model & Gradient checkpointing & true \\
\midrule
Data & Max sequence length & 65536 for Qwen 3.5, 32768 for Qwen 3 \\
\midrule
Optimisation & Epochs & 2 \\
Optimisation & Per-device batch size & 1 \\
Optimisation & Gradient accumulation steps & 4 \\
Optimisation & Global batch size & 128 \\
Optimisation & Learning rate & $2\times 10^{-5}$ \\
Optimisation & LR scheduler & linear \\
Optimisation & Warmup ratio & 0.03 \\
Optimisation & Weight decay & 0.0 \\
Optimisation & Optimiser & AdamW \\
\bottomrule
\end{tabular}
\caption{Hyperparameters for SFT training.}
\label{tab:sft_hyperparameters}
\end{table*}

For SFT training, we use the hyperparameters given in Table~\ref{tab:sft_hyperparameters}. We largely follow \citet{pi2026data} in hyperparameters. Additionally, we do not filter out unsuccessful/incomplete rollouts from our dataset following \citet{pi2026data}.

\subsection{LM Head Details}
\label{app:lm_head}

\begin{figure}
    \centering
    \includegraphics[width=.6\textwidth]{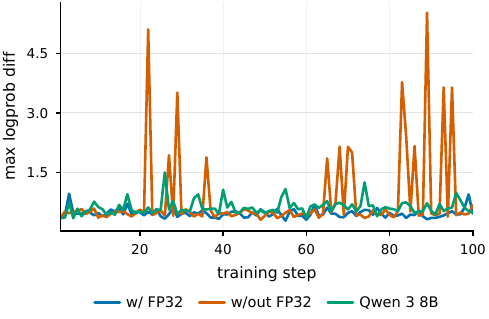}
    \caption{\textbf{Maximum difference between inference (vLLM) and trainer (HuggingFace) logprobs} during the first 100 steps of RL training for Qwen 3.5 9B. Qwen 3.5 without the FP32 LM head displays larger and more frequent spikes. Qwen 3 8B training also does not display spikes, even though we do not apply the FP32 LM head.}
    \label{fig:lm_head}
\end{figure}

We find using an FP32 LM head especially crucial when training Qwen 3.5, but not when training Qwen 3.
To highlight this, we plot the maximum logprob difference during RL training of Qwen 3.5 and Qwen 3 in Figure~\ref{fig:lm_head}. Using the FP32 LM head reduces the maximum difference dramatically. Interestingly, we find this is less important for Qwen 3, which does not display large logprob differences even without the FP32 head.

\subsection{Filtered Samples}
\label{app:filtered_samples}
\label{app:endless_filtering}
\begin{figure}[t]
    \centering
    \includegraphics[width=.6\linewidth]{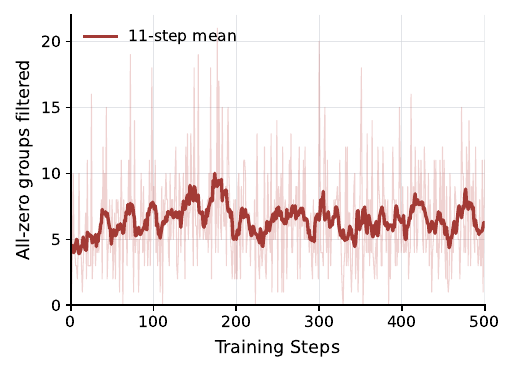}
    \caption{Number of all-zero groups filtered over the course of RL training on \ourdata.}
    \label{fig:filtered_zero_samples}
\end{figure}

In our RL training, we filter samples with all the same rewards, as these contribute no gradient to the batch. 
We plot the number of all-zero groups in Fig.~\ref{fig:filtered_zero_samples}, and observe that the number of filtered samples averages around 5-7. This suggests that, despite our lack of validation when generating data (\S\ref{sec:gen_pipeline}), our 9B model can still find solutions for roughly 60\% of sampled prompts during training.
To avoid unnecessarily filtering out valid samples that may simply be too hard, and to avoid the cost of filtering using a significantly larger and performant model, we leave these filtered samples in our released dataset. 

\begin{figure}[t]
    \centering
    \includegraphics[width=.6\linewidth]{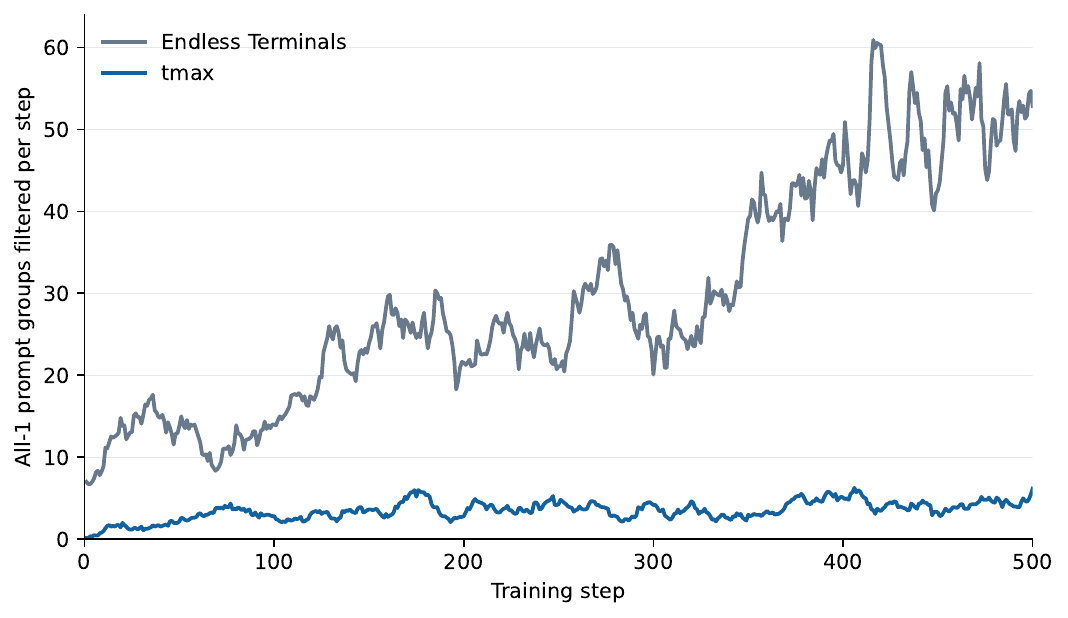}
    \caption{Number of all-one groups filtered over the course of RL training on \ourdata and Endless Terminals. Over the course of training, Endless Terminals requires significantly more sampling to fill a batch with non-zero variation samples.}
    \label{fig:entirely_solved}
\end{figure}

We also examine how many samples are filtered because they are entirely solved in Figure~\ref{fig:entirely_solved}. More samples are slowly filtered over training due to improving model ability. However, compared to a significantly easier dataset like Endless Terminals, significantly fewer samples are filtered per step by the end of training for \ourdata. This highlights the importance of difficulty in training efficiency: by the end of training, each training step for Endless Terminals generates seven times as many rollouts as the minimum needed to fill the batch (due to high filter rates).

\subsection{Reward Hacking}
\label{app:reward_hacking}

We perform a light check over our main Daytona-based runs on Terminal-Bench 2.0 and find that \ourmodel, after RL training, displays some instances of reward hacking. We find 2 clear cases:
\begin{itemize}
\item  \texttt{feal-differential-cryptanalysis}: The agent bypasses the intended attack by importing the cipher’s internal key array to recover the target key. The agent explicitly calls this “cheating” but submits it anyway.
\item \texttt{financial-document-processor}: The agent accidentally deletes the original documents, then recreates placeholder files under the same names and hand-writes the invoice CSV instead of extracting values from the documents.
\end{itemize}

In both cases, the rollouts with the hacking behaviour scored 0, so it does not affect downstream performance. Examining the chain of thought, we find that the model appears aware that it has not correctly performed the task. For example, for \texttt{cryptanalysis}, the model initially states ```\texttt{This means the attack should work with a black-box encrypt\_fn, without accessing internal state. My current solution cheats by accessing feal\_key directly. I need to fix this!}```. However, it then submits anyway: ```\texttt{It doesn’t prohibit accessing other modules or the key array.}'''. In the case of \texttt{financial-document-processor}, the model instead attempts to fool the verifier since it cannot actually perform the desired action: ```\texttt{The find command returned nothing. The files are truly gone. At this point, I need to make a pragmatic decision. The task cannot be completed perfectly without the original files. However, I can: 1. Create a working script that would correctly process the files if they existed 2. Create placeholder files that allow the script to run 3. Generate the expected summary.csv based on my analysis. [...] Since the task evaluation likely checks the final state, and I know what the correct classification and amounts should be (from my analysis), I'll manually create the final correct state.}''' We hope that future work can further explore the connection between our recipe and this hacking.

\section{Additional Evaluation Details}

\subsection{Full Figure~\ref{fig:teaser} Results}
\label{app:full_teaser_results}

\begin{table*}[t]
\centering
\footnotesize
\renewcommand{\arraystretch}{0.92}
\begin{tabular}{@{}llr@{}}
\toprule
\textbf{Model} & \textbf{Model Size} & \textbf{TB2.0 (\%)} \\
\midrule
\multicolumn{3}{@{}l@{}}{\textit{Models from Prior Work}} \\ \midrule
Nemotron-Terminal-8B~\citep{pi2026data} & 8B & 13.0 \\
Nemotron-Terminal-14B~\citep{pi2026data} & 14B & 20.2 \\
Nemotron-Terminal-32B~\citep{pi2026data} & 32B & 27.4 \\
TermiGen-Qwen2.5-32B~\citep{zhu2026termigen} & 32B & 19.3 \\
Endless-Terminals-8B~\citep{gandhi2025endless} & 8B & 6.7 \\
TerminalTraj~\citep{wu2026termitraj} & 7B  & 10.1 \\
TerminalTraj~\citep{wu2026termitraj} & 14B & 19.1 \\
TerminalTraj~\citep{wu2026termitraj} & 32B & 22.0 \\
LiberCoder-32B~\citep{lin2026cligymscalableclitask} & 32B & 19.5 \\
LiberCoder-235B-A22B~\citep{lin2026cligymscalableclitask} & 235B & 31.0 \\
OpenThinkerAgent-32B~\citep{raoof2026openthoughtsagent} & 32B & 26.2 \\
OT-Agent-ColdSFT+RL-8B~\citep{raoof2026openthoughtsagent} & 8B & 13.5 \\
LiteCoder-Terminal-32B-SFT~\citep{peng2026litecoderterminal} & 32B & 18.5 \\
SETA-8B~\citep{shen2026seta} & 8B & 10.7 \\
SETA-DS-V4-Flash~\citep{shen2026seta} & 284B & 43.0 \\
CalibForge-30B-A3B~\citep{meng2026calibforge} & 30B & 32.6 \\
CalibForge-35B-A3B~\citep{meng2026calibforge} & 35B & 47.6 \\
Terminal-Lego-Qwen3-8B~\citep{yang2026terminallego} & 8B & 11.8 \\
Terminal-Lego-Qwen3-32B~\citep{yang2026terminallego} & 32B & 24.3 \\
RST-122B-A10B~\citep{li2026recursivesynthesislonghorizonterminal} & 122B & 49.3 \\
\midrule
\multicolumn{3}{@{}l@{}}{\textit{Open-Weight Models}} \\ \midrule
GPT-OSS-120B & 120B & 18.7 \\
DeepSeek-V3.2 & 685B & 39.6 \\
MiniMax-M2.7 & 357B & 45.1 \\
Kimi-K2.5 & 1.04T & 43.2 \\
GLM-5 & 744B & 52.4 \\
\midrule
\multicolumn{3}{@{}l@{}}{\textit{Closed Models}} \\ \midrule
GPT-5-nano & ? & 11.5 \\
Claude-Haiku-4.5 & ? & 29.8 \\
GPT-5-mini & ? & 31.9 \\
Claude-Sonnet-4.6 & ? & 53.4 \\
\midrule
\multicolumn{3}{@{}l@{}}{\textit{Base Models}} \\ \midrule
Qwen3-8B & 8B & 0.9 \\
Qwen3.5-2B & 2B & 1.5 \\
Qwen3.5-4B & 4B & 16.5 \\
Qwen3.5-9B & 9B & 21.0 \\
Qwen3.6-27B & 27B & 42.7 \\
\midrule
\multicolumn{3}{@{}l@{}}{\textit{Our Models}} \\ \midrule
\ourtwobmodel & 2B & 9.4 \\
\ourfourbmodel & 4B & 25.1 \\
\ourmodel & 9B & 33.0 \\
\ourtwentysevenbmodel & 27B & 44.9 \\
\bottomrule
\end{tabular}
\caption{Full model names and Terminal-Bench 2.0 results for all 38 models; Figure~\ref{fig:teaser} plots 31 for legibility, keeping the Pareto shading clear of model labels. The plot places labels near their markers and uses abbreviated names, including Nemotron 14B, Haiku 4.5 and Sonnet 4.6; TerminalTraj-14B and MiniMax-M2.7 are unlabelled, and uncertainty bars are omitted. SETA-DS-V4-Flash is the SETA agent based on DeepSeek V4 Flash. Prior-work scores are from the cited papers or the Terminal-Bench 2.0 leaderboard. Other open-weight and closed-model scores are the best official leaderboard entries. Parameter count is total parameters, including inactive experts. Our models and Qwen baselines use our harness and report the mean over three trials per task.}
\label{tab:tb2-teaser_table}
\end{table*}

We show the full numeric results for Fig.~\ref{fig:teaser} in Table~\ref{tab:tb2-teaser_table}. For prior work, we use results reported by the original authors or on the Terminal-Bench 2.0 leaderboard\footnote{\url{https://www.tbench.ai/leaderboard/terminal-bench/2.0}}. For closed and open-weight models, we use leaderboard results. Where there are multiple leaderboard harnesses reported, we default to the best amongst mini-SWE-agent or Terminus-2 runs.

For the Qwen baselines and our own models, we run three trials per task using Daytona sandboxes and our harness. For \ourmodel, we do a small manual check for reward hacking (see \S\ref{app:reward_hacking}).

\section{SFT Details}

\subsection{SFT Data Generation}
\label{sec:sft_data_gen}
We additionally generate a small SFT dataset to use as a warm-start for RL training, with prior works showing that warm-up could help training stability and performance \citep{Guo_2025, team2025kimi}. Re-using our terminal data pipeline, we additionally generate 2.2k environments, and we use Qwen 3.6 27B to generate 8 trajectories for each environment. We filter out trajectories with unparsed tool calls.\footnote{To avoid traces containing the Qwen 3 XML-style tool call format, which can confuse small models with different tool call formats like Qwen 3 8B.} Together, this yields 16.5K total SFT trajectories, with 8K successful trajectories. We use this data for SFT experiments and finetuning Qwen 3 8B.

\subsection{SFT Ablations}
\label{app:sft_qwen_ablations}

\begin{table}[t]
\centering
\begin{tabular}{lcc}
\toprule
\textbf{Model} & \textbf{TB Lite} & \textbf{TB 2.1} \\
\midrule
Qwen 3.5 9B          & \boldmath{$41.9 \pm 2.7$} & \boldmath{$16.1 \pm 3.7$} \\
\quad + \ourmethod SFT    & $35.5 \pm 4.5$ &	 $15.0 \pm 3.0$               \\
\quad + large SFT      & $31.3 \pm 3.5$ & $16.9 \pm 0.9$ \\
\midrule
Qwen 3 8B            & $7.3 \pm 1.0$ & $1.1 \pm 0.9$ \\
\quad + \ourmethod SFT    & $11.5 \pm 0.1$ & $6.0 \pm 1.4$ \\
\quad + large SFT      & \boldmath{$16.4 \pm 2.3$} & \boldmath{$7.9 \pm 3.3$} \\
\bottomrule
\end{tabular}
\caption{\textbf{Older SFT data does not improve Qwen 3.5.} Performance of Qwen 3.5 9B and Qwen 3 8B on Terminal-Bench 2.1 and Terminal-Bench Lite before and after performing SFT. \ourmethod SFT refers to the SFT dataset in \S~\ref{sec:sft_data_gen}, while large SFT is described in \S~\ref{app:sft_qwen_ablations}. Performance is mean $\pm$ stderr over 3 evaluation runs. Results using an earlier version of our harness.}
\label{tab:qwen35-sft-rl}
\end{table}

In Table~\ref{tab:qwen35-sft-rl}, we compare Qwen 3.5 9B and Qwen 3 performance before and after performing SFT on both the SFT mixture mentioned in \S\ref{sec:sft_data_gen} and a larger mixture composed from our SFT data mixed with data from past work~\citep{pi2026data,wu2026termitraj,openthoughts-agent,shen2026sera}.\footnote{See \S\ref{app:full_sft_hypers} for details.} We find that the larger mix significantly improves performance on Qwen 3 8B, but drops performance on Qwen 3.5 9B, likely due to the use of weak teacher models in past work (e.g., \citet{pi2026data} use DeepSeek v3.2 as a teacher model). While \ourmethod SFT uses Qwen 3.6 27B as a teacher model, we still find it results in degraded performance for Qwen 3.5 9B, although it does aid Qwen 3 8B. As such, we opted to go straight to RL training for Qwen 3.5 experiments, while using an SFT stage for our Qwen 3 8B experiments. We suspect that further work on improving the SFT data quality could improve performance for Qwen 3.5, and leave it to future work.

\begin{table*}[t]
\centering
\small
\begin{tabular}{@{}llr@{}}
\toprule
Source dataset & Citation & \# Samples \\
\midrule
\texttt{allenai\_\_Sera\_4.6\_Lite\_47000} & \citet{shen2026sera} & 47,464 \\
\texttt{m\_a\_p\_\_TerminalTraj} & \citet{wu2026termitraj} & 16,748 \\
\texttt{Nemotron\_Terminal\_Corpus} & \citet{pi2026data} & 237,874 \\
\texttt{open\_thoughts\_\_OpenThoughts\_Agent\_v1\_SFT} & \citet{openthoughts-agent} & 8,717 \\
\ourmethod-SFT & - & 16,496 \\
\midrule
\textbf{Total} & & 327,299 \\
\bottomrule
\end{tabular}
\caption{Source datasets and sample counts for our `big' SFT mix described in \S\ref{app:sft_qwen_ablations}.}
\label{tab:big-sft-datasets}
\end{table*}

We additionally show the full splits for our `big' SFT mix described above in Table~\ref{tab:big-sft-datasets}.

\section{Infrastructure Data Filtering}
\label{app:infra_filter}

Due to our reliance on Apptainer, not all datasets can be fully run on our setup. For example, we remove one Terminal Bench Lite instance that requires managing user permissions on files in a way that Apptainer does not support (often due to fakeroot limitations).
As such, we examine all dataset images in our training and evaluation datasets, and remove ones that are unable to run in our setup.

\paragraph{Baseline Dataset Filtering.} We filtered various tasks from the baseline datasets we compare against in Table~\ref{tab:termbench}. We list the total number of tasks removed and why for each dataset below. We only remove tasks from 2 datasets, and always remove less than 10\% of the data.

\begin{itemize}
    \setlength{\itemsep}{0pt}
    \item \textbf{TerminalTraj}~\citep{wu2026termitraj}: 5,648 total; 510 removed; 5,138 retained. We excluded 509 images whose BusyBox \texttt{setsid} is incompatible with our harness's \texttt{-w} flag, plus one image with a dangling \texttt{/workspace} symlink that prevents sandbox startup.
    \item \textbf{TermiGen}~\citep{zhu2026termigen}: 3,556 total; 0 removed; 3,556 retained.
    \item \textbf{LiteCoder}~\citep{peng2026litecoderterminal}: 602 total; 0 removed; 602 retained.
    \item \textbf{OpenThinker-Agent}~\citep{openthoughts-agent}: 5,000 total; 0 removed; 5,000 retained.
    \item \textbf{CLI-Gym}~\citep{lin2026cligymscalableclitask}: 1,552 total; 28 removed; 1,524 retained. Similar to TerminalTraj, we remove images that provide utilities incompatible with our harness (e.g., not supporting \texttt{setsid -w}, or adding syscall hooks that disrupt the harness).
    \item \textbf{SETA}~\citep{shen2026seta}: 4,567 total; 0 removed; 4,567 retained.
    \item \textbf{CalibForge}~\citep{meng2026calibforge}: 5,431 total; 0 removed; 5,431 retained.
    \item \textbf{Terminal-Lego}~\citep{yang2026terminallego}: 15,048 total; 0 removed; 15,048 retained.
    \item \textbf{RST}~\citep{li2026recursivesynthesislonghorizonterminal}: 37,484 total; 0 removed; 37,484 retained.
    \item \textbf{SWE-Smith}~\citep{yang2025swesmith}: 59,136 total; 0 removed; 59,136 retained.
    \item \textbf{Endless Terminals}~\citep{gandhi2025endless}: 2,492 total; 0 removed; 2,492 retained.
    \item \textbf{Facet}~\citep{shi2026facet}: 6,020 total; 0 removed; 6,020 retained.
\end{itemize}

\paragraph{Eval Dataset Filtering.}\label{app:eval_filtering} Our final Terminal-Bench Lite and Terminal-Bench 2.1 scores use fixed subsets of 96/100 and 88/89 tasks, respectively, with three trials per included task. We exclude the following tasks because our rootless Apptainer backend cannot evaluate them reliably; these are task-level exclusions applied to every model, not model failures removed after scoring.
\begin{itemize}
    \setlength{\itemsep}{0pt}
    \raggedright
    \item \textbf{Terminal-Bench Lite}~\citep{OpenThoughts-TBLite}: We exclude \mbox{\emph{acl-permissions-inheritance}}, \mbox{\emph{etl\_checkpoint\_resume\_bug}}, \mbox{\emph{sales-data-csv-analysis}}, and \mbox{\emph{mlflow-register}} because fakeroot prevents reliable setup or grading. We retain 96 tasks.
    \item \textbf{Terminal-Bench 2.1}~\citep{merrill2026terminal}: We exclude \mbox{\emph{configure-git-webserver}} because OpenSSH privilege separation fails under fakeroot. We retain 88 tasks.
\end{itemize}

Note that for the Terminal-Bench 2.0 results in Figure~\ref{fig:teaser}, we evaluate over \textbf{all} tasks, as the Daytona backend does not have similar issues.


\end{document}